# Deep Learning and Glaucoma Specialists: The Relative Importance of Optic Disc Features to Predict Glaucoma Referral in Fundus Photos

Sonia Phene, BS[1*]
R. Carter Dunn, MS, MBA[1*]
Naama Hammel, MD,[1*]
Yun Liu, PhD[1]
Jonathan Krause, PhD[1]
Naho Kitade, BA[1]
Mike Schaekermann, BS[1]
Rory Sayres, PhD[1]
Derek J. Wu, BS[1]
Ashish Bora, MS[1]
Christopher Semturs, MS,[1]
Anita Misra, BTech[1]
Abigail E. Huang, MD[1]
Arielle Spitze, MD[2,3]
Felipe A. Medeiros, MD, PhD[4]
April Y. Maa, MD[5,6]
Monica Gandhi, MD[7]
Greg S. Corrado, PhD[1]
Lily Peng, MD, PhD[1**]
Dale R. Webster, PhD[1**]

*Equal contribution
**Equal contribution

Affiliations:
[1]Google Health, Google LLC, Mountain View, CA, USA
[2]Virginia Ophthalmology Associates, Norfolk, VA, USA
[3]Department of Ophthalmology, Eastern Virginia Medical School, Norfolk, VA
[4]Department of Ophthalmology, Duke University, Durham, NC, USA
[5]Department of Ophthalmology, Emory University School of Medicine, Atlanta, GA, USA
[6]Ophthalmology Section, Atlanta Veterans Affairs Medical Center, Atlanta, GA, USA
[7]Dr. Shroff's Charity Eye Hospital, New Delhi, India

Corresponding Author:
Naama Hammel, MD
Google AI Healthcare, Google LLC
1600 Amphitheatre Pkwy, Mountain View, CA 94043
[nhammel@google.com](nhammel@google.com)

Funding: Google funded this study and had a role in its approval for publication.

Conflict of interest: SP, RCD, NH, YL, JK, NK, MS, RS, DJW, AB, CS, AM, AEH, AS, GSC, LP, DRW were employed by or consultants for Google for the duration of this study. AYM and MG have no competing interests to declare. FAM: Consultant for Carl-Zeiss Meditec, Inc; Reichert, Inc., Alllergan, Novartis, Quark Pharmaceuticals, Stealth Biotherapeutics, Galimedix Therapeutics, Inc.; Research support from Carl-Zeiss Meditec, Heidelberg Engineering, Reichert, Dyopsys, Inc.

**This paper has been accepted and appears in *Ophthalmology***

## Abstract

***Purpose:*** Develop and validate a deep learning (DL) algorithm that predicts referable glaucomatous optic neuropathy (GON) and optic nerve head (ONH) features from color fundus images, determine the relative importance of these features in referral decisions by glaucoma specialists and the algorithm, and compare the performance of the algorithm with eye-care providers.

***Design:*** Development and validation of a DL algorithm.

***Participants:*** Fundus images from screening programs, studies, and a glaucoma clinic.

***Methods:*** A DL algorithm was trained using a retrospective dataset of 86,618 images, assessed for glaucomatous ONH features and referable GON (defined as ONH appearance worrisome enough to justify referral for comprehensive examination) by a panel of 43 graders. The algorithm was validated using 3 datasets: dataset "A" (1,205 images, 1 image/patient; 18.1% referable) - images adjudicated by panels of glaucoma specialists; dataset "B" (9,642 images, 1 image/patient; 9.2% referable) - images from a diabetic teleretinal screening program; and dataset "C" (346 images, 1 image/patient; 81.7% referable) - images from a glaucoma clinic.

***Main outcome measures:*** The algorithm was evaluated using the area-under-the-receiver-operating-characteristic-curve (AUC), sensitivity, and specificity for referable GON and glaucomatous ONH features.

***Results:*** The algorithm's AUC for referable GON was 0.945 (95% CI, 0.929-0.960) in dataset "A", 0.855 (95% CI, 0.841-0.870) in dataset "B", and 0.881 (95% CI, 0.838-0.918) in dataset "C". Algorithm AUCs ranged between 0.661-0.973 for glaucomatous ONH features. The algorithm had significantly higher sensitivity than 7/10 independent graders not involved in determining the reference standard, including 2/3 glaucoma specialists, and had higher specificity than 3 graders (including 1 glaucoma specialist), while remaining comparable to the others. For both glaucoma specialists and the algorithm, the most crucial features related to referable GON were the presence of vertical cup-to-disc ratio ≥ 0.7, neuroretinal rim notching, retinal nerve fiber layer defect, and bared circumlinear vessels.

***Conclusions:*** A DL algorithm trained on fundus images alone can detect referable GON with higher sensitivity and comparable specificity to eye care providers. The algorithm maintained good performance on an independent dataset with diagnoses based on a full glaucoma workup.

## Introduction

Glaucoma, the "silent thief of sight", is the leading cause of preventable, irreversible blindness world-wide.[1,2] Glaucoma currently affects about 70 million individuals, and is projected to affect about 112 million in 2040.[2] The disease can remain asymptomatic until severe, and an estimated 50%-90% of people with glaucoma remain undiagnosed.[3–5] Thus, glaucoma screening is recommended for early detection and treatment.[6,7] However, this currently requires a clinical exam combined with quantitative functional and structural measurements, accessible only to a small percentage of the world's population. A cost-effective tool to detect glaucoma could expand screening access to a much larger patient population. Such a tool is currently unavailable.[8]

Retinal fundus photography is a well-established diagnostic tool for eye diseases, enabling fundus images to be evaluated for the presence of retinal and optic nerve head (ONH) pathologies.[9,10,8] Glaucoma, characterized by progressive degeneration of retinal ganglion cells and loss of their axons, results in characteristic structural ONH and retinal nerve fiber layer (RNFL) changes.[11] These changes can be detected in fundus images[12] and are the most important aspect of glaucoma diagnosis.[13]

Physical features that have been associated with glaucomatous optic neuropathy (GON) include neuroretinal rim thinning and/or notching, increased or asymmetric cup-to-disc ratio (CDR), excavation of the cup, RNFL thinning, disc hemorrhages, parapapillary atrophy (PPA), laminar dots, nasalization of central ONH vessels, and baring of circumlinear vessels.[14,15] However, several of the features listed may appear in otherwise healthy optic nerves,[16,17] may result from non-glaucomatous pathology,[18,19,20] or are poorly defined in the literature.[21] The relative importance of these features for the diagnosis of glaucoma has not been validated, and practitioners may weigh features differently, often focusing on increased CDR, neuroretinal rim thinning and notching, disc hemorrhages, and RNFL defects.[22]

Deep learning (DL)[23] has been applied to produce highly accurate algorithms that can detect eye conditions such as diabetic retinopathy (DR) with accuracy comparable with that of human experts.[24–26] The use of this technology may aid screening efforts, and recent work demonstrates value in assisting ophthalmologists.[27] DL algorithms have also been developed for other diseases, including glaucoma.[26,28–31] However, the diagnosis of glaucoma poses several challenges that distinguish it from conditions such as DR. While DR has well-established clinical guidelines for diagnosis from fundus images, the diagnosis of glaucoma consists of the compilation of clinical exam data with functional and structural testing. There is no single agreed-upon set of guidelines for glaucoma detection from fundus images.[32] Teleretinal screening programs have worked around these limitations by focusing on subsets of the potential diagnostic features present in fundus images. However, these criteria vary between programs and the relationship of the individual features to overall glaucoma assessment is not well characterized. This strongly limits the clinical value of these diagnoses, and correspondingly limits the value of DL algorithms developed around these heuristics.

The present work aims to extend the utility of DL algorithms by explicitly bridging the gap between individual diagnostic features present in fundus imagery and referable GON assessment by experts. We develop an algorithm that is trained on both individual pathologies, as well as overall GON referability. Our study aims to answer three main questions: Can we develop a DL algorithm that performs at or above expert level in referable GON detection? How do glaucoma specialists weigh the relative importance of individual features towards a referable GON decision? Do DL algorithms consider the same features with similar relative importance?

## Methods

**Datasets**

Development datasets

The development datasets for this study consisted of fundus images obtained from multiple sources: EyePACS,[33] Inoveon,[34] and the Age Related Eye Disease Study (AREDS)[35] in the United States, three eye hospitals in India (Aravind Eye Hospital, Sankara Nethralaya, and Narayana Nethralaya), and the UK Biobank.[36]

Validation datasets

Three datasets were used for validation. Validation dataset "A" comprised a random subset of color fundus images from EyePACS, Inoveon, UK Biobank, AREDS, and Sankara Nethralaya. Validation dataset "B" comprised macula-centered color fundus images from the Atlanta Veterans Affairs (VA) Eye Clinic diabetic teleretinal screening program. Validation dataset "C" comprised macula-centered and disc-centered color fundus images from the glaucoma clinic at Dr. Shroff's Charity Eye Hospital, New Delhi, India.

All images were de-identified according to HIPAA Safe Harbor before transfer to the study investigators. Ethics review and Institutional Review Board exemption was obtained using Quorum Review Institutional Review Board. From each source, only one image per patient was used in the final analysis, and each image appeared only in either the development or validation sets (see details in "Development of the Algorithm" below).

**Development Dataset Grading**

Gradability

Graders assessed the gradability of specific ONH features in every image. Gradability was measured using the degree to which each distinct feature was detected in the image, and the ability of each feature to be identified (based on image quality, blurring, media opacity, or any other confounding reason). In addition, graders were asked to assess gradability for determining referable GON. If graders selected "ungradable" for a particular feature or referable GON, then no grade was collected for that aspect.

Feature-Level Grading

The American Academy of Ophthalmology's Primary Open-Angle Glaucoma Suspect Preferred Practice Patterns provides a list of features that may indicate GON: vertical elongation of the optic cup, excavation of the cup, thinning of the RNFL, notching of the neuroretinal rim, thinning of the inferior superior neuroretinal rim or both, disc hemorrhage, parapapillary atrophy, nasalization of central ONH vessels, baring of circumlinear vessels, and absence of neuroretinal rim pallor. Additionally, the finding of a laminar dot sign (visible lamina cribrosa fenestrations in the cup) is viewed by some clinicians as a sign of glaucomatous damage to the ONH.[37,38] Violation of the Inferior > Superior > Nasal > Temporal (ISNT) neuroretinal rim area rule is viewed by some clinicians as a sign of glaucomatous rim loss. ISNT rule variants, focusing on the inferior, superior and temporal quadrants also have been validated.[39] To enable systematic training of graders, we aggregated these features (**Table S1**), developed grading guidelines for each, and iterated on the guidelines with a panel of three fellowship-trained glaucoma specialists to increase inter-rater agreement.

Referable GON grading

In addition to these features, we also developed guidelines for a four-point GON assessment (as shown in **Table S1**) where the "high-risk glaucoma suspect" or "likely glaucoma" levels were considered referable, that is, the ONH appearance was worrisome enough to justify referral for comprehensive examination. Graders were asked to provide a referable GON grade after evaluating the image for each of the features on the list.

<u>Graders</u>

A total of 43 graders (14 fellowship-trained glaucoma specialists, 26 ophthalmologists, and 3 optometrists) were trained on the grading guidelines and were required to pass a certification test before grading the development or validation "A" datasets.

**Development of the Algorithm**

Data preprocessing, algorithm design, and hyperparameters are detailed in the Supplement. Using this data, a deep convolutional neural network with the Inception-v3 architecture[40] was developed and trained in TensorFlow.[41] To speed up training, the network was initialized using parameters from a network pre-trained to classify objects in the ImageNet dataset[42]. This procedure is similar to that previously described by Krause et al.[25]

The input to the network is a single color fundus photograph resized to 587x587 resolution. The network was trained to output a probability for each of the possible values for each feature. For instance, for referable GON, the network outputs four probabilities corresponding to no, low-risk, high-risk, and likely glaucoma. These probabilities sum to one. The network is able to make multiple predictions simultaneously, such as referable GON and the presence or absence of various ONH features.

The development datasets (see **Table 1**) consisted of two disjoint subsets: training and tuning. The training set of 86,618 images was used to optimize the network parameters. This dataset was enriched for "referable GON" images using active learning,[43] a machine learning technique, to preferentially increase the number of relevant examples. The tuning set consisted of 1,508 images, each independently graded by three glaucoma specialists (from a group of 11 specialists), and was used to determine thresholds for three operating points for the algorithm: high sensitivity, high specificity, and a balanced sensitivity and specificity point (see Supplement for details). The tuning dataset and validation dataset "A" (described below) were randomly chosen from a pool that was also enriched, but using preliminary "screening" reviews by a separate panel of graders for images suspicious for a glaucomatous-appearing disc.

An ensemble of 10 networks[44] were trained on the same training set, and the outputs were averaged to yield the final prediction. Each network was built in an identical manner (see "Algorithm Design" in Supplement for details.

**Clinical Validation Datasets**

Three datasets were used for validation. For validation dataset "A" (1,205 images), the reference standard was determined by a rotating panel of 3 fellowship-trained glaucoma specialists (from a cohort of 12). Each fundus image was first independently graded by each glaucoma specialist in the panel ("round 1"). To resolve any disagreements between graders and thus reduce potential inadvertent grading errors and increase the quality of the reference standard, the same 3 glaucoma specialists (in a random order) reviewed the image a second time ("round 2"), this time with access to the annotations and comments from round 1 and previous graders in round 2. Grader identities were anonymized throughout the grading process. Each image was reviewed a maximum of 6 times (2 from each grader), or fewer if consensus was reached before the 6th review. By the completion of round 2, 49.3% (594 images) were fully adjudicated on all features and referable GON.

For images unresolved after round 2, the reference standard for gradability, each ONH feature, and referable GON was determined by the median of the 3 grades in round 2. The median corresponds to the majority vote if at least 2 graders agreed, and the "intermediate" grade if all 3 disagreed. This method was determined to closely approximate the full adjudication process described in Krause et al.[25] based on a comparison of multiple methods over 100 images described in **Tables S2** and **S3**. Images for which the final grade was "ungradable" were excluded from further analysis. Finally, both to reflect the fact that referral decisions are "binary" (yes/no), and to facilitate comparisons between validation datasets with different types of glaucoma assessment, we binarized both the model predictions and the reference standard. Specifically for validation dataset "A", "High-risk glaucoma suspect" or "likely glaucoma" levels were considered referable and ONH features were binarized according to **Table 2**.

To compare the performance of the algorithm with graders, a random subset of 411 images from validation dataset "A" was graded by 10 graders (3 fellowship-trained glaucoma specialists, 4 ophthalmologists, and 3 optometrists). None of these graders participated in creating the reference standard for those images. Assessments of graders were binarized in the same way as the reference standard described above.

For further evaluation on an independent population, the algorithm was also applied to a second validation dataset, dataset "B", composed of 9,642 macula-centered color fundus images from the Atlanta VA Eye Clinic diabetic teleretinal screening program. Glaucoma-related International Classification of Diseases (ICD) codes on a per-patient basis and ONH referral codes recorded in the patient's medical history at any point in time before the capture of the image or up to 1 year after the image was taken were used as a reference standard for this dataset. Presence of such a code was considered "referable" because these are patients that were either referred for an in-person exam by the diabetic teleretinal screening program or were given such a code during an in-person exam. Ophthalmologists at the VA screening program recorded the ICD codes with access to 3 fundus images per eye and the patient's clinical record. The specific codes used were ICD-9-CM codes for glaucoma (category 365) and ICD-10-CM codes for glaucoma (category H40). When a patient had an optic nerve head referral code on a particular visit, the image from that visit was chosen. When a patient with a glaucoma ICD code underwent multiple visits, the visit closest to the recorded date of the glaucoma-related ICD code was chosen. To evaluate our algorithm's image-level predictions against the patient-level codes, we applied the algorithm to the fundus images from that visit, typically one per eye. The image that produced the highest referable GON prediction was used in the final patient-level analysis; only one eye per patient was used in the final analysis.

Lastly, the algorithm was applied to validation dataset "C", comprising 346 macula-centered and disc-centered fundus images from the glaucoma clinic at Dr. Shroff's Charity Eye Hospital. The reference standard for this dataset was an eye-level diagnosis of glaucoma as determined by a glaucoma specialist based on a full glaucoma workup that included history, clinical exam, visual field, and optical coherence tomography (OCT) assessment. For the final evaluation, we randomly included one eye per patient and selected an image from that eye at random. The reference standard for these images consisted of eye level diagnoses on a three-tiered scale of "Normal", "Suspect", and "Glaucoma". A diagnosis of "Suspect" or "Glaucoma" was considered "referable" for purposes of comparison against the algorithm as these patients were undergoing regular follow-up.

**Evaluating the Algorithm**

The algorithm's performance was evaluated on each of the three validation datasets. In all cases, the algorithm only takes as input and makes predictions based on the fundus photograph although the reference standard for that photo may be determined using other clinical data as described previously.

The prediction for referable GON was a continuous number between 0 and 1, corresponding to the predicted likelihood of that condition being present in the image. Receiver operating characteristic (ROC) curves were plotted by varying the operating threshold.

**Evaluating ONH feature importance**

To assess which ONH features were associated most strongly with referable GON, a multivariable logistic regression analysis was carried out to compute the odds ratio and beta coefficients (i.e., natural logarithms of odds ratios) for each feature on referable GON. Five logistic regression analyses were carried out. For validation dataset "A", analyses were carried out using the reference standard, the round 1 median (see "Clinical Validation Datasets" section above), and the algorithm's predictions (see **Table 3**). Analyses were also carried out on the training dataset median and the tuning dataset median (see **Table S4**).

**Statistical Analysis**

To compute the confidence intervals (CIs) for the algorithm's AUC, we used a non-parametric bootstrap procedure[45] with 2,000 samples. CIs for sensitivities and specificities were calculated using the exact Clopper-Pearson interval.[46] To compare the algorithm's performance (sensitivity and specificity) with that of graders, we used the two-tailed McNemar test.[45] To compare the distributions of vertical CDR, we used the Kolmogorov-Smirnov test.[47] We measured inter-grader agreement using a weighted Krippendorff's alpha, which is robust to multiple graders.[48] The weighting function was the difference between grades (e.g. the distance between the first grade (non-glaucomatous) and third grade (high-risk glaucoma suspect) is 3-1=2).

# Results

A DL algorithm was developed using a training set of 86,618 fundus images and tuning set of 1,508 fundus images that were assessed for the presence of glaucomatous ONH features and referable GON. The images were graded by a panel of 43 graders using a set of detailed grading guidelines (**Table S1**). Patient demographics and image characteristics for the development and validation sets are summarized in **Table 1**.

We first evaluated the algorithm on validation dataset "A" in which referable GON was assessed by glaucoma specialists, using the images alone, and achieved an AUC of 0.945 (95% CI, 0.930-0.959), with sensitivity of 80.0% and specificity of 90.2% at the balanced operating point (**Figure 1A**). **Figure 2** summarizes the performance of the algorithm in detecting referable GON on a subset of 411 images from validation dataset "A". On this random subset, the algorithm achieved an AUC of 0.933 (95% CI, 0.899-0.961) with sensitivity of 76.1% and specificity of 92.3% at the balanced operating point. 10 eye care providers who were not involved in creating the reference standard graded this same subset. The performance of these 10 graders on this same subset is plotted for comparison on the ROC curve for algorithm performance (**Figure 2, Table S4**).

Graders' sensitivities ranged from 29.2-73.6% (37.5-63.9% for glaucoma specialists, 29.2-73.6% for ophthalmologists and 29.2-70.8% for optometrists) and specificities ranged from 75.8-92.6% (85.5-90.6% for glaucoma specialists, 75.8-90.6% for ophthalmologists and 87.6-92.6% for optometrists). The algorithm was significantly more sensitive than 7 of 10 graders (including 2 of 3 glaucoma specialists) with no statistically significant difference for the other 3 graders. The algorithm showed significantly higher specificity than one glaucoma specialist and two ophthalmologists (**Table S4**) and no statistically significant difference in specificity for the other 7 of 10 graders.

Next, we evaluated the performance of the algorithm relative to real clinical decisions as the reference standard. First, for validation dataset "B", in which ophthalmologists made referral decisions with access to 3 fundus images and clinical data, the algorithm achieved an AUC of 0.855 (95% CI, 0.841-0.870) (**Figure 1B**). Second, on validation dataset "C", which had glaucoma and glaucoma suspect diagnoses based on a full clinical workup including history, clinical exam, VF, and OCT, the algorithm achieved an AUC of 0.881 (95% CI, 0.836-0.918) (**Figure 1C**).

Additionally, we looked at the performance of the algorithm on binarized ONH features (**Table 2**). Our algorithm showed good performance in predicting the presence of the different ONH features, with AUCs ranging between 0.661 and 0.973. We also evaluated the relative importance of the different ONH features in the predictions of the algorithm and graders (**Table 3, Figure S1A-B**). A vertical CDR≥0.7, presence of a neuroretinal rim notch, presence of a RNFL defect, and presence of bared circumlinear vessels showed the strongest correlation with an overall assessment of referable GON for both glaucoma specialists and the algorithm. Presence of a neuroretinal rim notch, presence of laminar dots, and presence of a disc hemorrhage (in descending order) showed a stronger correlation with the referable GON of the reference standard labels compared with the round 1 majority grades. See **Table S4** for similar analyses on the training and tuning datasets.

To better understand the association of ONH features with referable GON assessment, we plotted the distributions of ONH features in the refer and no-refer categories (**Figures 3D, S2**). Most strikingly, the vertical CDR distributions were significantly different between the refer and no-refer categories ($p<0.001$ for differences in the distribution) (**Figure 3A**). We further plotted referral rate as a function of each ONH feature's grade, mimicking a "positive predictive value" analysis of each feature. Marked increases in referral rates were observed when RNFL defect, disc hemorrhage, laminar dot sign and beta PPA were present or possibly present. **Figure 4A** includes fundus examples in which the algorithm's referral predictions differed from the reference standard in validation dataset "B." **Figure 4B** includes fundus examples where the algorithm's referral predictions differed from the reference standard in validation dataset "C".

We further sliced the results for validation dataset "A" based on self-reported sex: the algorithm achieved an AUC of 0.939 (95% CI, 0.916-0.960) for females (n=571, **Figure S3A**), and an AUC of 0.947 (95% CI, 0.921-0.968) for males (n=504, **Figure S3B**). Validation dataset "B" included too few females (4.7%) to make the analysis by self-reported sex meaningful, and self-reported sex was not available for validation dataset "C" at the time of study.

## Discussion

In the present study, we developed a DL algorithm for the prediction of referable GON and the prediction of glaucomatous ONH features from fundus images. We validated this algorithm on multiple independent datasets from different patient populations, using established adjudication methods, clinical referral decisions, and actual clinical diagnoses as reference standards. In addition, we analyzed the relative importance that glaucoma specialists attribute to the different ONH features in their overall assessment of referable GON, and whether the DL algorithm learned a similar relationship.

**Performance of the algorithm**

Our DL algorithm achieved an AUC of 0.945 for the detection of referable GON using a reference standard based on glaucoma specialists' assessment of fundus photographs alone, applied to an independent set of patients from the same populations sampled for the development dataset (validation dataset "A"). When evaluated against the referral decisions of eye-care providers in a teleretinal screening program (validation dataset "B"), our algorithm achieved an AUC of 0.855. This decrease in performance relative to validation dataset "A" may be explained by the difference in reference standard and patient populations between the two validation datasets. Specifically, while both the algorithm and graders of validation dataset "A" only had access to a single fundus image from each patient, graders for validation dataset "B" had access to the electronic medical record, including patient history and previous notes containing clinical data. When evaluated on a validation dataset with a reference standard of glaucoma diagnosis based on a full glaucoma workup (validation dataset "C"), the model achieved an AUC of 0.881. Similar to validation dataset "B", this decrease in performance relative to validation dataset "A" may be explained by differences in reference standard and patient population. Nonetheless, the fact that our algorithm, which was developed using eye care providers' grades based on fundus photographs alone, maintained good performance on an independent dataset with grades based on a full workup, suggests that the algorithm's prediction of referable GON is fairly well correlated with the results of a full glaucoma workup.

Several previous studies reported performance of DL algorithms with AUCs ranging between 0.91-0.986.[26,28–31] Our work advances the field by demonstrating performance on par with or surpassing eye-care providers after addressing the following limitations of prior works: (1) grading of images by non-glaucoma specialists, limiting the validity of GON identification;[26,29,31] (2) lack of a consistent definition of glaucoma for the reference standard, limiting the effectiveness of algorithm performance evaluation;[28] (3) exclusion of low quality images from the training set, limiting the usefulness of the algorithm in real-life settings;[30] (4) exclusion of images on which graders had disagreed from the validation set, skewing the dataset towards easier cases;[30] and (5) use of fundus images zoomed-in on the optic nerve, limiting assessment of the RNFL.[31]

**ONH Features analysis**

We demonstrate herein that a DL algorithm trained only on fundus photographs containing ONHs can accurately predict glaucoma assessments made by glaucoma specialists with access to additional patient information (validation set "C"). This suggests that combining information about different ONH pathologic features has clinically relevant predictive power for overall glaucoma risk. Therefore, we wanted to quantitatively assess how glaucoma specialists make clinical decisions on suspicious ONHs. For each image, we asked our graders to indicate the presence or absence of certain glaucoma-associated ONH features, in addition to overall assessment of referable GON. It is worth noting that for referral decision, graders were guided to use their "gut feel" tied to the clinical actions they deemed necessary, rather than directly map features to overall risk (see **Table S1**). We have also trained an algorithm for the detection of most known glaucomatous ONH features, showing good performance in predicting the presence of the different ONH features, with AUCs ranging between 0.661-0.973.

Not surprisingly, we found that the presence of a vertical CDR≥0.7 was highly correlated with an overall assessment of an image as referable GON. Of several additional ONH features listed as possibly indicative of GON in the AAO PPP,[14] the presence of a neuroretinal rim notch, RNFL defect, or a bared circumlinear vessel were most correlated with referable GON, with similar ranking order and coefficients for algorithm predictions and reference standard (as shown in **Table 3**).

Studies training DL algorithms for glaucoma detection based on color fundus images used slightly varying sets of ONH features to determine glaucoma risk, with different thresholds for CDR. Li et al.[29] defined "referable GON" as any of the following: CDR>0.7, rim thinning or notching, disc hemorrhage or RNFL defect. Shibata et al.[30] labelled images as "glaucoma" according to the following: focal rim notching or generalized rim thinning, large CDR with cup excavation with or without laminar dot sign, RNFL defects with edges at the optic nerve head margin, disc hemorrhages, and peripapillary atrophy. Ting et al.[26] identified "possible glaucoma" if any of the following was present: CDR≥0.8, focal thinning or notching of the neuroretinal rim, disc hemorrhages, or RNFL defect. Other studies[28,31] did not report image grading guidelines.

Our model found the ONH features of CDR, notch, and RNFL defect to be most correlated with referable GON, which is in line with the approaches described previously.

Interestingly, features mostly not assessed in previous studies, that is baring of circumlinear vessels, laminar dots (somewhat assessed in Shibata et al.) and nasalization of central ONH vessels, were significantly correlated with referable GON, whereas disc hemorrhage was not. The relatively low ranking of disc hemorrhage in our study may be explained by the study population. Images used in this study were largely derived from DR screening programs. We asked our graders to mark the presence of any disc hemorrhage, regardless of its presumed etiology. Therefore, some of the disc hemorrhages identified by our graders may have been due to DR.

Our study is the only study to specifically ask graders to assess for the presence of baring of circumlinear vessels (an ONH feature first described by Herschler and Osher in 1980[49]) and nasalization of central ONH vessels (a feature included in the AAO's PPP), both associated with glaucomatous optic nerves, yet rarely delineated separately by glaucoma specialists in clinical practice. Interestingly, the presence of bared circumlinear vessels was among the ONH features that were most highly correlated with referable GON, whereas nasalization of central ONH vessels was not. One reason for this distinction may be that the presence of bared circumlinear vessels is associated with adjacent rim thinning, making it more likely to be correlated with GON than nasalization of central ONH vessels, often seen in large discs.[50]

Our findings may be helpful in deciphering the decision making process by which glaucoma specialists identify an ONH as glaucomatous, often previously described by many as a "gut feeling". Understanding which features are weighed most highly by glaucoma specialists for grading an ONH as more likely to be glaucomatous may help in the development of training programs to enable non-specialists to recognize those features in a tele-ophthalmology or screening setting. Similarly, by training the algorithm to detect individual ONH features, we may be able to better explain what the algorithm's predictions rely on, and thus gain insight into what is frequently referred to as a "black box" of machine learning.

**Limitations and future work**

Our study has several limitations. First, although ONH assessment for detection of referable glaucoma risk is commonly accepted in screening and teleophthalmology settings, it is well established that subjective evaluation of optic disc photographs suffers from low reproducibility, even when performed by expert graders[51,52] (e.g. a panel of glaucoma specialists, as carried out in this study). Second, the diagnosis of glaucoma is not based on ONH appearance alone, but also relies on the compilation of risk factors (such as race, age, and family history), and repeated clinical, functional and structural testing over time, which were included in only one validation dataset in our study. In addition, our study did not provide graders with absolute disc size or subjects' race, two important factors in the assessment of an ONH for glaucoma.[53,54] Thus, the predictions of our algorithm are not diagnostic for glaucoma, but rather are an assessment of the ONH appearance for referable GON. Nonetheless, we believe that our algorithm's generalization to an external dataset with glaucoma diagnoses based on a full glaucoma work-up demonstrates the robustness of our algorithm for flagging suspicious discs and detection of referable GON, which can be useful in screening settings. A longitudinal dataset consisting of fundus images, clinical examination data, OCT scans and VF tests could be used to train an algorithm that may be able to diagnose glaucoma, and perhaps even predict progression.

Regarding leveraging other imaging modalities, Spectral-Domain (SD) OCT has become a wide-spread tool for glaucoma diagnosis and follow-up over the last decade, enabling visualization and quantitative analyses of the RNFL and ganglion cell layers.[55] Recent works applying DL to SD and Swept-Source OCT images were able to detect early glaucoma and distinguish healthy from mild glaucoma with high accuracies.[56,57] However, although three dimensional SD-OCT scans enable better structural analysis of the ONH than two-dimensional color fundus photographs, optic disc fundus photography is the least expensive and globally the most commonly used imaging modality for the structural assessment of the ONH. Fundus photography is also widely deployed in DR screening programs where algorithms to detect risk of non-DR pathology such as age related macular degeneration and glaucoma could be deployed. Thus, there will remain a role for fundus photography as an imaging modality for screening purposes, especially in low resource settings.

In conclusion, we developed a DL algorithm with higher sensitivity and comparable specificity to eye care providers in detecting referable GON in color fundus images. The algorithm's prediction of referable GON maintained good performance on an independent dataset with diagnoses based on a full glaucoma workup. Additionally, our work provides insight into which ONH features drive GON assessment by glaucoma specialists. These insights may help improve clinical decisions for referring patients to glaucoma specialists based on ONH findings during diabetic fundus image assessments. We believe that an algorithm such as this may also enable effective screening for glaucoma in settings where clinicians trained to interpret ONH features are not available, thus reaching underserved populations world-wide. The use of such a tool presents an opportunity to reduce the number of undiagnosed patients with glaucoma, and thus provides the chance to intervene before permanent vision loss occurs.

## Acknowledgements

This work would not have been possible without the assistance of the following institutions that graciously provided de-identified data: Dr. Shroff's Charity Eye Hospital, Atlanta Veterans Affairs Medical Center, Inoveon, Aravind Eye Hospital, Sankara Nethralaya, and Narayana Nethralaya. From Google AI Healthcare, we'd like to thank William Chen, BA, Quang Duong, PhD, Xiang Ji, MS, Jess Yoshimi, BS, Cristhian Cruz, MS, Olga Kanzheleva, MS, Miles Hutson, BS, and Brian Basham, BS for their software infrastructure contributions. We'd also like to thank Jorge Cuadros, OD, PhD, from EyePACS for data access and helpful conversations. This research has been conducted using the UK Biobank Resource under Application Number 17643.
Some images used for the analyses described in this manuscript were obtained from the NEI Study of Age-Related Macular Degeneration (NEI-AMD) Database found at [https://www.ncbi.nlm.nih.gov/projects/gap/cgi-bin/study.cgi?study_id=phs000001.v3.p1] through dbGaP accession number [phs000001.v3.p1.c1]. Funding support for NEI-AMD was provided by the National Eye Institute (N01-EY-0-2127). We would like to thank NEI-AMD participants and the NEI-AMD Research Group for their valuable contribution to this research.

# Tables and figures

| | Development Datasets* | | Validation Datasets | | |
|---|---|---|---|---|---|
| **Characteristics** | **Training dataset** | **Tuning Dataset** | **Validation Dataset A** | **Validation Dataset B (VA Atlanta)** | **Validation Dataset C (Dr. Shroff)** |
| No. of images (1 image per patient) | 86,618 | 1,508 | 1,205 | 9,642 | 346 |
| No. of graders | 43 graders: 14 glaucoma specialists; 26 ophthalmologists; 3 optometrists | 11 glaucoma specialists | 12 glaucoma specialists | 6 ophthalmologists | 4 glaucoma specialists |
| No. of grades per image | 1-2 | 3 | 3-6 | 1 | 1 |
| Grades per grader, median (IQR) | 861 (225-2952) | 388 (96-553) | 185 (93-731) | N/A** | NA** |
| Data used to define the reference standard | 45 degree fundus photograph | 45 degree fundus photograph | 45 degree fundus photograph | Health factors and ICD codes | Full glaucoma workup including VF and OCT |
| **Patient demographics** | | | | | |
| No. of patients | 86,618 | 1,508 | 1,205 | 9,642 | 346 |
| Age, median (IQR); no. of images for which age was available | 57 (49-65), 64,861 | 57 (49-64.8); 1,386 | 57 (49.5-64); 1,115 | 64 (68.7-57.5); 9,642 | N/A** |
| Females, No./total of images for which sex was known (%) | 32,413/62,178 (52.1) | 731/1,349 (54.2) | 571/1,075 (53.2) | 458/9,642 (4.7) | N/A** |
| **Glaucoma/GON gradability distribution** | | | | | |

Table 1. Baseline Characteristics

| Images gradable for glaucoma, No./total (%) among images for which glaucoma gradability was assessed | 74,973/86,127 (87.0) | 1,451/1,508 (96.2) | 1,171/1,204 (97.3) | 9,642/9,642 (100)*** | 346/346 (100)*** |
|---|---|---|---|---|---|
| **Glaucoma/GON risk distribution** | | | | | |
| No. non-Glaucomatous (%) | 35,877 (47.8) | 849 (57.1) | 687 (57.5) | 8,753 (90.8) | 63 (18.2) |
| No. low-risk glaucoma suspect (%) | 20,740 (27.6) | 259 (17.4) | 290 (24.3) | N/A**** | N/A**** |
| No. high-risk glaucoma suspect (%) | 13,180 (17.5) | 268 (18.0) | 170 (14.1) | N/A**** | 175 (50.6) |
| No. likely glaucoma (%) | 5,307 (7.1) | 110 (7.4) | 48 (4.0) | 890 (9.2) | 108 (31.2) |
| No. "Referable glaucoma" (%) | 18,487 (24.6) | 378 (25.4) | 218 (18.1) | 890 (9.2) | 283 (81.7) |

Abbreviations: VA, Veterans Affairs; IQR, Interquartile Range; OCT, optical coherence tomography; VF, visual field; GON, Glaucomatous Optic Neuropathy.

* Prevalence of referable GON risk images is higher than in the general population in part due to active learning, a machine learning technique used to preferentially increase the number of relevant examples (methods).

** Data not available.

*** All images in validation sets B and C were of overall "adequate image quality", but not specifically labeled by graders for glaucoma gradability.

**** Finer-grained categorization not available, see Methods.

Table 2. Logistic Regression Models to Understand the Relative Importance of Individual Optic Nerve Head (ONH) Features for Glaucomatous Optic Neuropathy (GON) Referral Decisions in Validation Dataset A: The Reference Standard, Algorithm Predictions, and Round 1 Majority (N=1015)*

| | Reference Standard | | | Algorithm Predictions | | | Round 1 Majority | | |
|---|---|---|---|---|---|---|---|---|---|
| | Odds Ratio | p-value | Rank | Odds Ratio | p-value | Rank | Odds Ratio | p-value | Rank |
| Vertical CD Ratio ≥ 0.7 | 581.671** | < 0.001 | 1 | 347.861 | < 0.001 | 1 | 475.757† | < 0.001 | 1 |
| Notch: Possible or Yes | 29.438 | < 0.001 | 2 | 9.564 | 0.021 | 3 | 4.158 | 0.218 | 4 |
| RNFL Defect: Possible or Yes | 10.740 | < 0.001 | 3 | 13.098 | < 0.001 | 2 | 12.946 | < 0.001 | 2 |
| Circumlinear Vessels: Present + Bared | 4.728 | < 0.001 | 4 | 6.241 | < 0.001 | 4 | 4.852 | < 0.001 | 3 |
| Laminar Dot: Possible or Yes | 3.594 | < 0.001 | 5 | 3.320 | < 0.001 | 7 | 3.882 | < 0.001 | 6 |
| Disc Hemorrhage: Possible or Yes | 3.221 | 0.043 | 6 | 2.178 | 0.369 | 9 | 1.649 | 0.508 | 9 |
| Nasalization Emerging: Yes | 3.162 | 0.008 | 7 | 4.253 | 0.001 | 5 | 4.014 | < 0.001 | 5 |
| Rim Comparison: I < S | 2.560 | 0.004 | 8 | 3.512 | < 0.001 | 6 | 3.033 | < 0.001 | 7 |
| Nasalization Directed: Yes | 2.230 | 0.002 | 9 | 3.010 | < 0.001 | 8 | 2.239 | 0.001 | 8 |
| Rim Comparison: S < T | 1.461 | 0.799 | 10 | 1.257 | 0.894 | 11 | 1.175 | 0.919 | 11 |
| Beta PPA: Possible or Yes | 1.319 | 0.226 | 11 | 1.584 | 0.076 | 10 | 1.357 | 0.192 | 10 |

Abbreviations: CD, Cup-to-Disc; RNFL, Retinal Nerve Fiber Layer; I, Inferior; S, Superior; PPA, ParaPapillary Atrophy; T, Temporal;

* Some images in Validation Dataset A were excluded from analysis due to being ungradable on referral criteria or for a specific ONH feature.

** Extreme odds ratios here indicate almost perfect correlation between the feature and the final referral prediction or grade.

| | AUC [95% CI] | Number of labeled images | Prevalence (%) | Binary cutoffs |
|---|---|---|---|---|
| Rim width I Vs. S | 0.661 [0.594-0.722] | 1,162 | 8.2 | I<S vs. I>S or I~=S |
| Rim width S Vs. T | 0.946 [0.897-0.981] | 1,156 | 1.6 | S<T vs. S>T or S~=T |
| Notch | 0.908 [0.852-0.956] | 1,162 | 2.6 | Yes/Possible vs. No |
| Laminar dot sign | 0.950 [0.937-0.963] | 1,013 | 24 | Yes/Possible vs. No |
| Nasalization emerging | 0.973 [0.954-0.987] | 1,166 | 4.7 | Yes vs. Possible/No |
| Nasalization directed | 0.957 [0.944-0.969] | 1,167 | 15.9 | Yes vs. Possible/No |
| Baring of circumlinear vessels | 0.723 [0.688-0.755] | 1,154 | 22.7 | Present and clearly bared vs. All else |
| Disc hemorrhage | 0.758 [0.666-0.844] | 1,173 | 2.1 | Yes/Possible vs. No |
| Beta PPA | 0.933 [0.914-0.948] | 1,170 | 16.9 | Yes/Possible vs. No |
| RNFL defect | 0.778 [0.706-0.843] | 973 | 6.5 | Yes/Possible vs. No |
| Vertical CD ratio | 0.922 [0.869-0.963] | 1,154 | 4.6 | ≥0.7 vs. <0.7 |

Table 3. Evaluation of Algorithm Performance for Detecting Presence of Individual Features on Validation Dataset A

Abbreviations: AUC, Area Under the Curve; CI, Confidence interval; I, Inferior; S, Superior; T, Temporal; PPA, ParaPapillary Atrophy; RNFL, Retinal Nerve Fiber Layer; CD, Cup-to-Disc

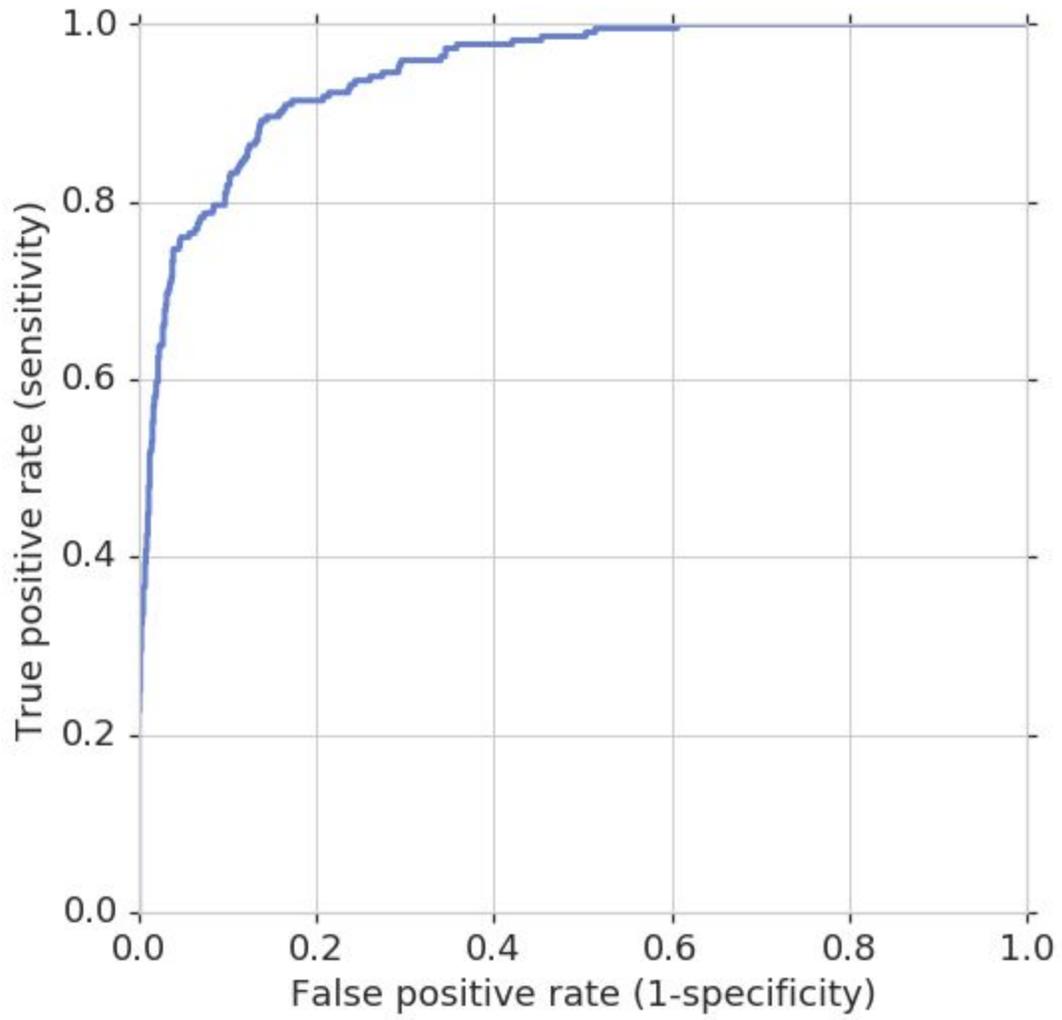

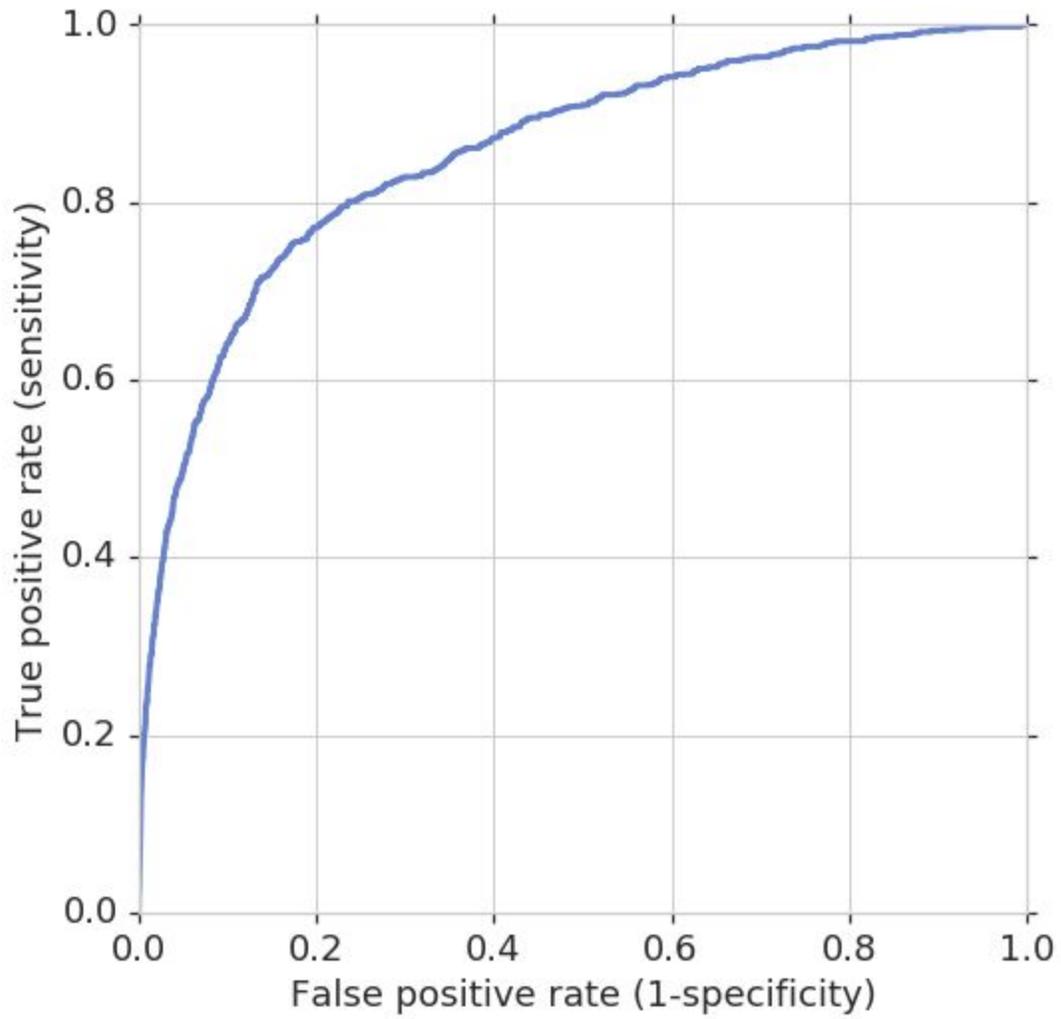

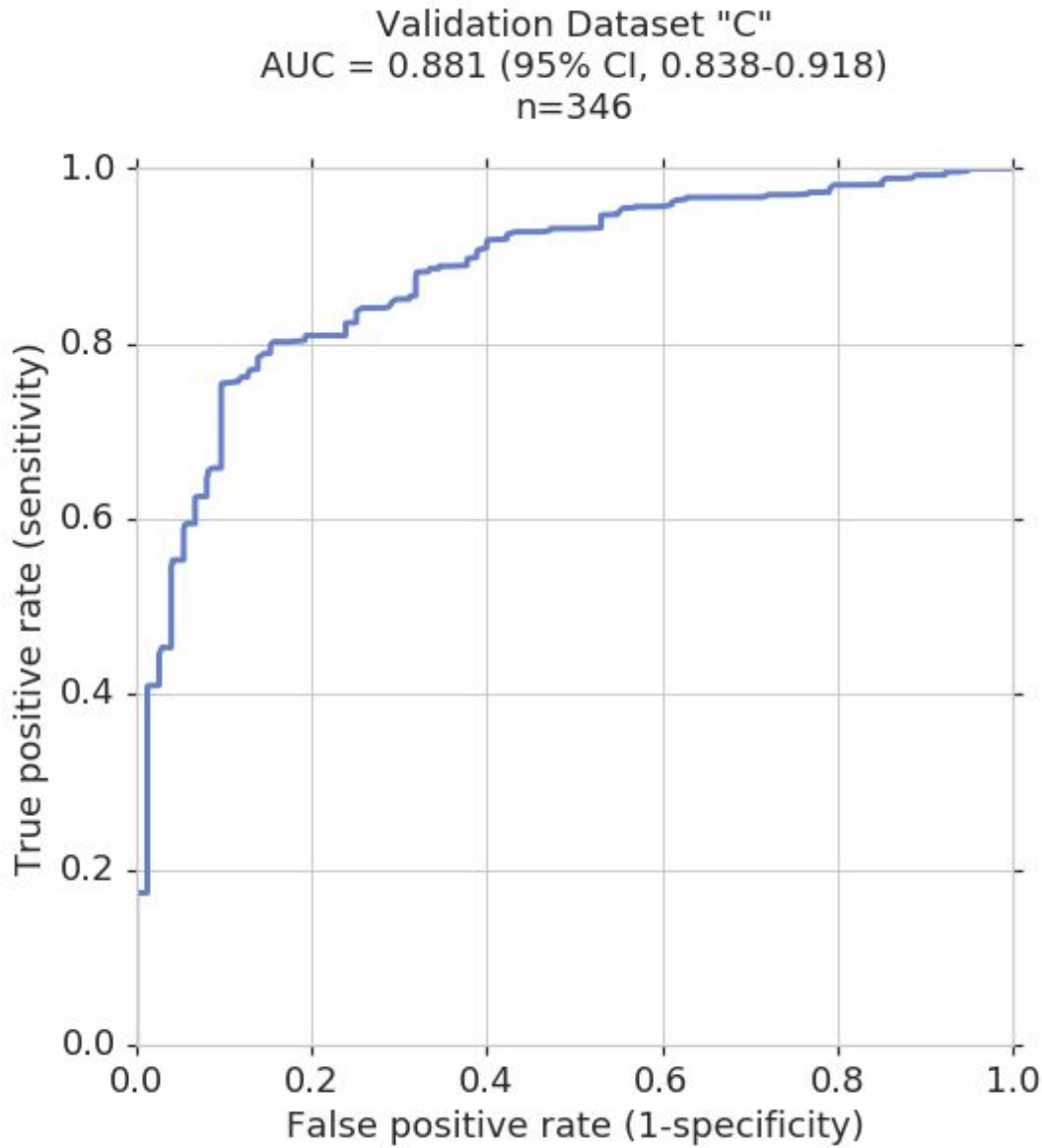

**Figure 1A-C**. **Receiver operating characteristic (ROC) curve analyses for referable Glaucomatous Optic Neuropathy (GON) risk in three independent validation datasets**
(A) Validation dataset "A", the reference standard for all images was determined by a panel of 3 glaucoma specialists. (B) Validation dataset "B" (VA Atlanta), the reference standard for all images was determined by glaucoma related ICD codes, assigned to images by eye-care providers at a screening program. (C) Validation dataset "C" (Dr. Shroff's Charity Eye Hospital), the reference standard was determined by glaucoma specialists based on full glaucoma workups.

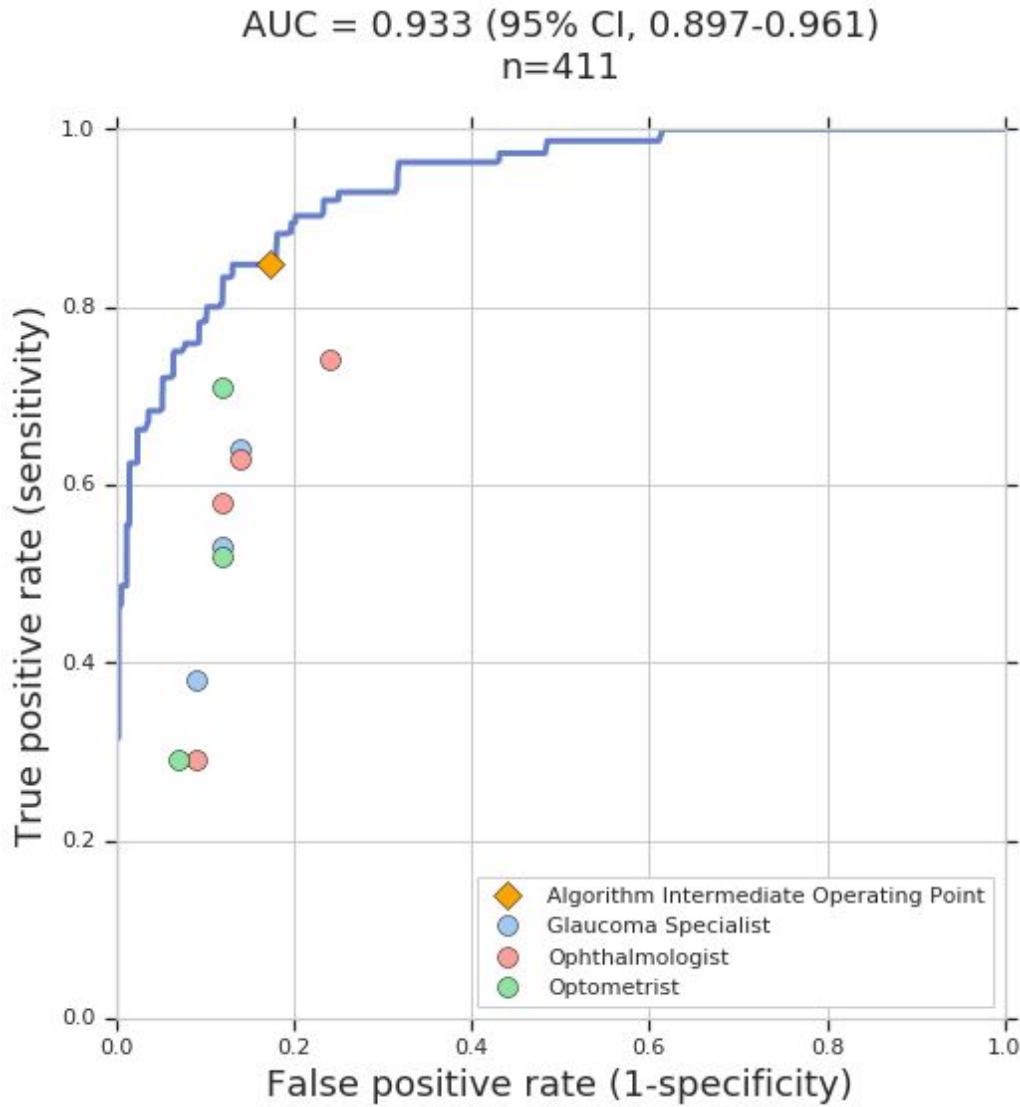

**Figure 2. Receiver operating characteristic (ROC) analysis for referable Glaucomatous Optic Neuropathy (GON) risk in a subset of validation dataset A (n=411) with comparison to clinicians.**
The algorithm is illustrated as a blue line, with 10 individual graders indicated by colored dots: glaucoma-specialists (blue), ophthalmologists (red), and optometrists (green). The diamond corresponds to the balanced operating point of the algorithm, chosen based on performance on the tuning set. For each image, the reference standard was determined by a different set of three glaucoma specialists in an adjudication panel (Methods). Images labeled by graders as 'ungradable' for glaucoma, were considered as 'refer' to enable comparison on the same set of images**.** For a sensitivity and specificity analysis excluding the 'ungradable' images on a per-grader basis, see **Table S5**. See **Figure 1A** for analysis on the entire validation dataset "A."

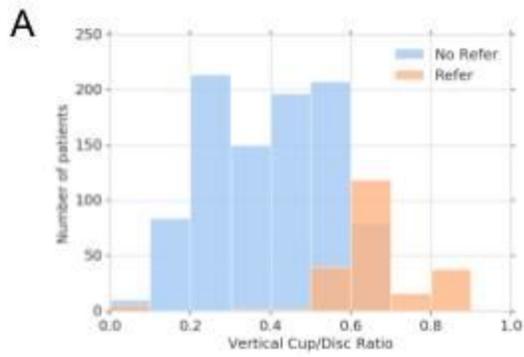
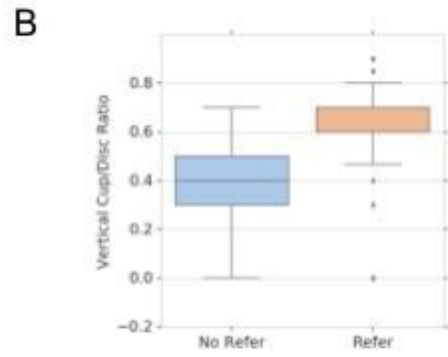
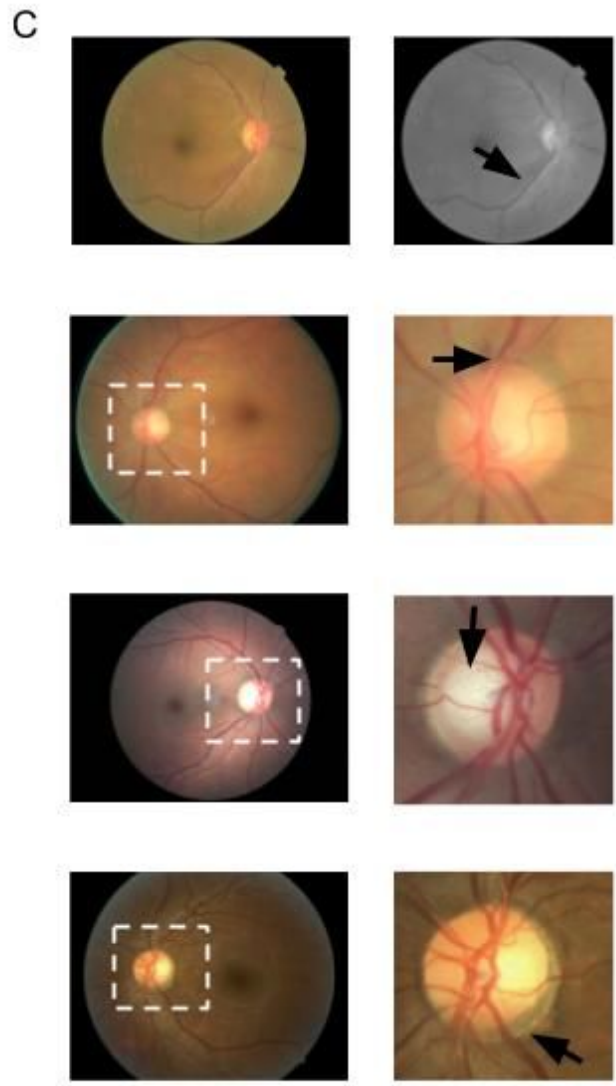
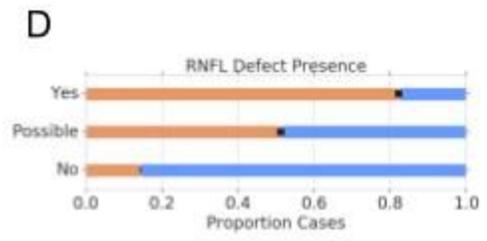
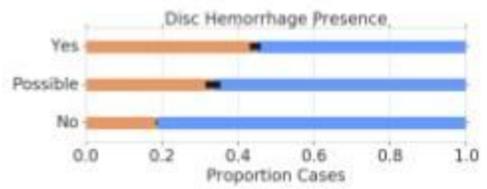
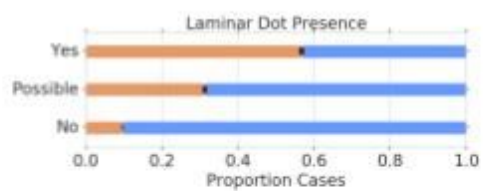
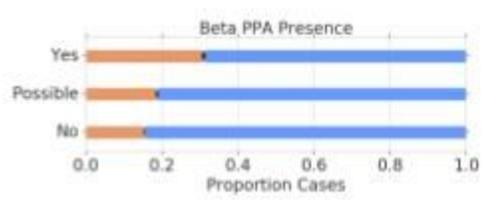

**Figure 3A-D**: **Proportions of selected Optic Nerve Head (ONH) feature grades amongst the refer/no-refer categories in validation data set "A"**.

**(A)** Distribution of images based on their vertical cup-to-disc-ratio, stratified by refer/no-refer categories of GON risk. **(B)** Box-plot of vertical cup-to-disc-ratio by refer/no-refer categories of GON risk. **(C)** ONH feature example images for RNFL defect, disc hemorrhage, laminar dot sign and beta PPA. **(D)** Corresponding distributions of ONH feature presence by refer/no-refer categories of GON risk. Error bars represent the 95% confidence intervals.

Abbreviations: GON, Glaucomatous Optic Neuropathy; ONH, Optic Nerve Head; RNFL, Retinal Nerve Fiber Layer; PPA, ParaPapillary Atrophy

| Validation Set "B": panel A ||
|---|---|
| **False Positive** | **False Positive** |
| 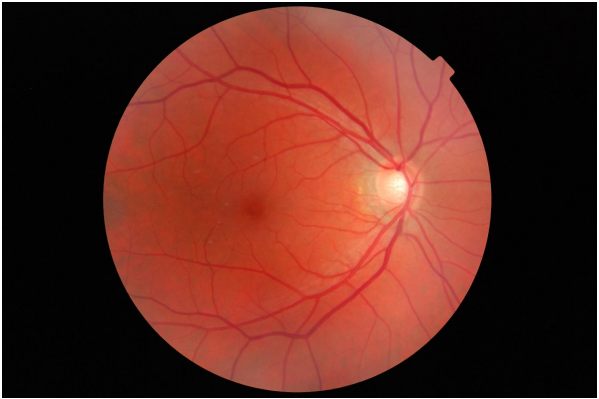 <br> Actual: No Refer <br> Predicted: Refer | 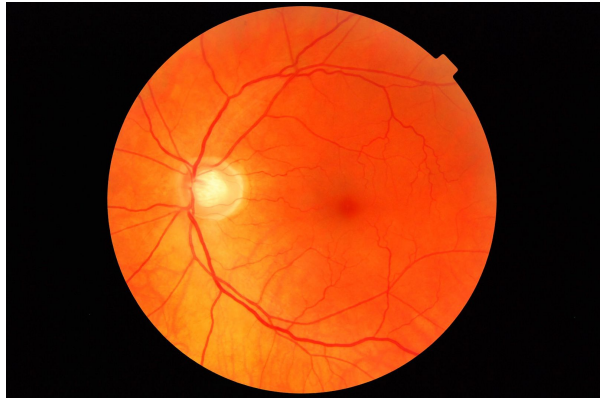 <br> Actual: No Refer <br> Predicted: Refer |
| **False Negative** | **False Negative** |
| 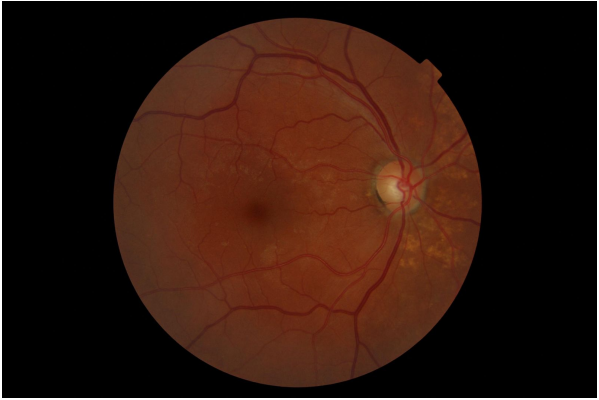 <br> Actual: Image taken 47 days after ICD glaucoma code <br> Predicted: No Refer | 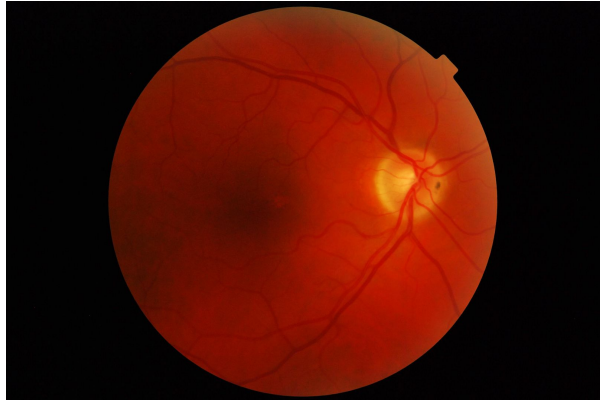 <br> Actual: Image taken 1603 days after ICD glaucoma suspect code <br> Predicted: No Refer |
| Validation Set "A": panel B ||
| **False Positive** | **False Negative** |

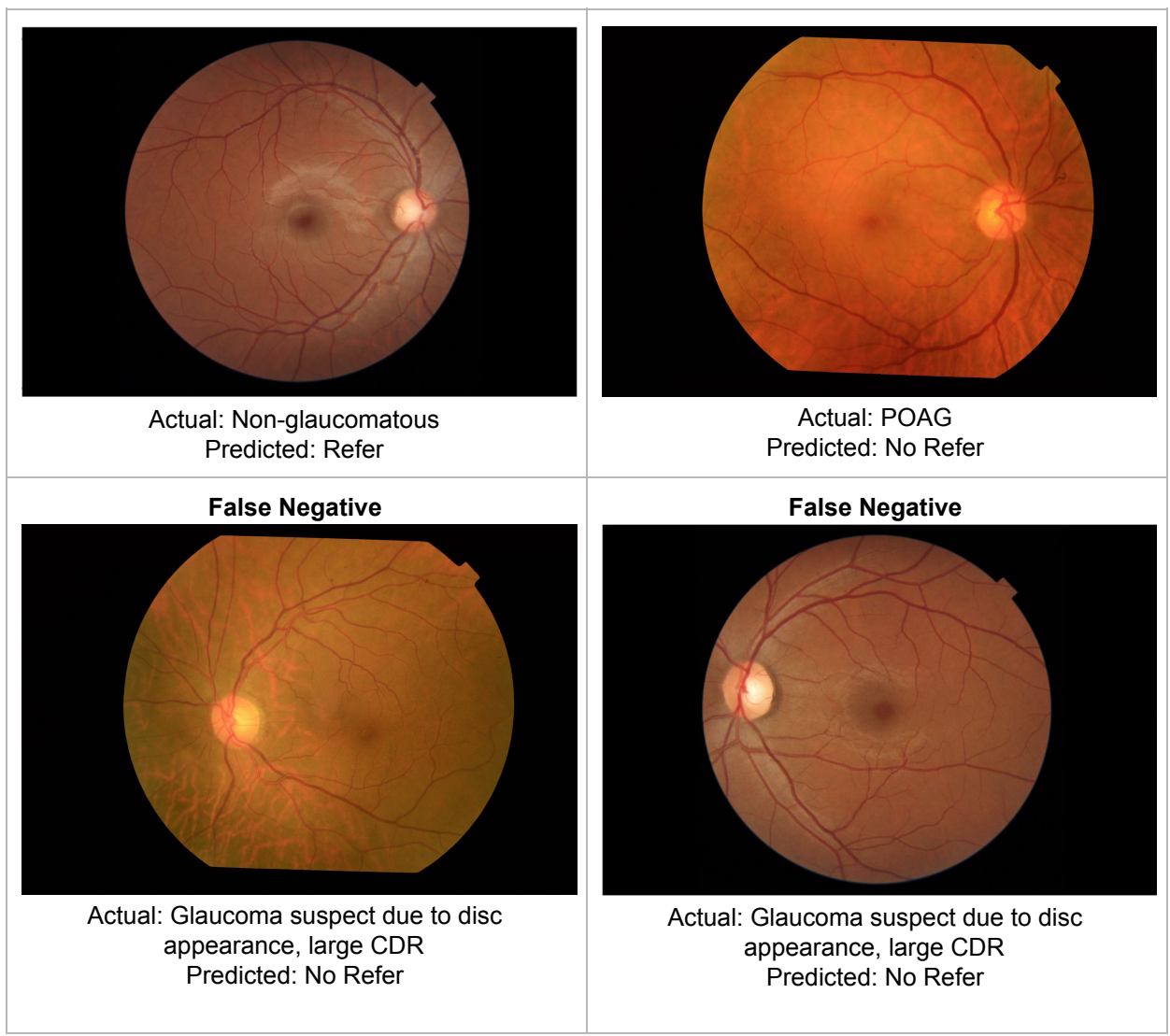

**Figure 4A-B. Examples of incorrect algorithm predictions on external validation datasets.**
A: Images had a referral prediction different than the reference standard for validation set "B". Discrepancies between ONH appearance and VA diabetic teleretinal screening program glaucoma referral decisions may be explained by VA providers' access to clinical data available in patients' electronic medical records.
B: Images that had a referral prediction different than the diagnosis reference standard for validation set "C". Discrepancies between model referral decisions vs. those based on clinical diagnoses may be explained by provider access to visual field and OCT findings as well as clinical data.
Abbreviations: ICD, International Classification of Diseases; ONH, Optic nerve head

# Supplement

| Table S1. Grading Guidelines |||
|---|---|---|
| | Question | Possible answers |
| Optic nerve head features ||| 
| Rim width assessment | Is the inferior rim (I) width greater than the superior rim (S) width? | <u>I clearly greater than S</u>: Inferior rim width is greater than superior rim width (the ratio I:S is 3:2 or more*)<br><u>Similar widths</u>: Inferior and superior rim widths are roughly the same<br><u>S clearly greater than I</u>: Superior rim width is greater than inferior rim width (the ratio S:I is 3:2 or more*) |
| | Is the superior rim (S) width greater than the temporal rim (T) width? | <u>S clearly greater than T</u>: Superior rim width is greater than temporal rim width (the ratio S:T is 3:2 or more*)<br><u>Similar widths</u>: Superior and temporal rim widths are roughly the same<br><u>T clearly greater than S</u>: Temporal rim width is greater than superior rim width (the ratio T:S is 3:2 or more*) |
| | Is there a notch in the neuroretinal rim? | <u>Yes</u>: A notch in the neuroretinal rim is present that: - has circumferential extent of up to 3 clock hours - has a change in curvature - does not fall only in the temporal quadrant<br><u>Possible</u>: Possible notch that is borderline on criteria for Yes<br><u>No</u>: No notch in the neuroretinal rim is present - Notch is entirely in the temporal quadrant (9 o'clock OD, 3 o'clock OS) should be marked as [No] |
| Laminar dot sign | Are laminar dots or striations visible? | <u>Yes</u>: Laminar dots/striations are visible within the cup<br><u>Subtle/Possible</u>: Possible laminar dots/striations visible within the cup<br><u>No</u>: No dots or striations present |
| Nasalization of central vascular trunk | Is central trunk emerging in the nasal third of the optic nerve head? | <u>Yes</u>: Central vascular trunk is emerging in nasal third of disc<br><u>Borderline</u>: Central trunk is emerging at the border of the central and nasal thirds, or Vein emerges in central third and artery emerges in nasal third<br><u>No</u>: Central trunk is emerging in the central or temporal thirds |
| | Is central vascular trunk directed nasally? | <u>Yes</u>: Central vascular trunk is directed nasally<br><u>minimally</u>: Central vascular trunk is minimally directed nasally |

| | | No: Central vascular trunk is not directed nasally |
|---|---|---|
| Baring of circumlinear vessels | Are bared circumlinear vessels present? | Present & Clearly not bared: Obvious circumlinear vessel/s are present and clearly none are bared<br>Present & Clearly bared: Obvious circumlinear vessel/s are present and clearly at least one is bared<br>Possibly present or possibly bared: circumlinear vessel possibly present or an obvious circumlinear vessel is possibly bared<br>No circumlinear vessels present: No circumlinear vessels are present |
| Disc hemorrhage | Is a hemorrhage at /near the disc present? | Yes: A hemorrhage at or near the disc is present that is within 2 vessel widths from the edge of the disc<br>Possible/Borderline: Heme that is borderline on criteria for Yes or possible heme that meets criteria<br>No: No hemorrhage at/near the neuroretinal rim |
| Beta PPA | Is beta PPA present? | Yes: Beta PPA is present (with or without alpha)<br>Possible: PPA is possibly present (e.g. atypical location, not sure if its alpha or Beta, etc.)<br>No: there is only alpha PPA, or there is no PPA at all |
| RNFL defect | Is an RNFL defect present? | Yes: RNFL defect is present that:<br>- follows an arcuate pattern<br>- extends all the way to the disc<br>- is wedge shaped, narrows as it gets closer to disc<br>or RNFL loss is such that only one clear sheen/RNFL absence border is seen<br>Possible: Possible RNFL defect that is borderline on criteria for Yes<br>No: No RNFL defect present |
| Vertical cup-to-disc ratio | Estimate the vertical cup-to-disc ratio to nearest 0.1 | 0.1-0.9 |
| Glaucoma risk | | |
| Risk assessment | Estimate risk for glaucoma | Non-glaucomatous: disc does not appear glaucomatous; No need for OCT / VF to rule out glaucoma; Follow up in about 2 years<br>Low-risk glaucoma suspect: Unlikely to have glaucoma but not completely normal; Order baseline VF and OCT if had the resources available; Follow up about 1 year<br>High-risk glaucoma suspect: Even chance to likely to have glaucoma; Order serial VFs and OCTs if had the resources available; Follow up about 4-6 months<br>Likely glaucoma: Almost certain to have glaucoma; Order serial VFs and OCTs if had the resources available; Probably needs treatment now |

| Other ONH findings | | |
|---|---|---|
| Assessment of ONH for presence of additional findings | Pallor out of proportion to cupping | Yes<br>No |
| | Disc swelling | Yes<br>No |
| | Optic nerve head tumor | Yes<br>No |
| | Melanocytoma | Yes<br>No |
| | Disc drusen | Yes<br>No |
| | Anomalous disc | Yes<br>No |
| | Other findings? | Free-text box |

Abbreviations: PPA, ParaPapillary Atrophy; RNFL, Retinal Nerve Fiber Layer; ONH, Optic Nerve Head; OCT, Optical Coherence Tomography; VF, Visual Field.

**Data preprocessing**

For algorithm training, input images were scale normalized by detecting the circular mask of the fundus image and resizing the diameter of the fundus to be 587 pixels wide. Images for which the circular mask could not be detected were not used in the development, tuning, or clinical validation sets.

Images from validation set B were all chosen from the macula-centered fundus field. If an "optic nerve head" referral code was present, then images from that visit were chosen. In cases where a glaucoma-related ICD code was present and there were multiple images, the image selected was the one with the date closest to the glaucoma ICD code date, up to one year after the ICD code was given.

**Algorithm design**

Our implementation very closely follows Krause et al.[25]

We used the following hyperparameters:
- Inception-v3 architecture
- Input image resolution: 587 x 587
- Learning rate: 0.001
- Batch size: 16
- Data augmentation:
    - Random horizontal reflections
    - Random vertical reflections

- Random brightness changes (with a max delta of 0.1147528) [see the TensorFlow function tf.image.random_brightness]
- Random saturation changes between 0.5597273 and 1.2748845 5 [see tf.image.random_saturation]
- Random hue changes (with a max delta of 0.0251488) [see tf.image.random_hue]
- Random contrast changes between 0.9996807 and 1.7704824 [see tf.image.random_constrast]

Each model in the 10-way ensemble was trained in this fashion for 250,000 steps. Experiments using more than 10 models did not yield any improvements on the tuning dataset. Model evaluations performed using an exponential moving average of parameters, with a decay factor of 0.9999.

**Tuning Dataset**

In algorithm development, the performance on the tuning set was used to select the algorithm checkpoint that yielded the highest AUC for referable GON risk. Once a checkpoint was determined, operating points at high sensitivity, high specificity, and a balanced point were chosen, and the thresholds at these values were then applied during evaluation on the validation sets. The balanced point was chosen to be the point where the absolute difference between sensitivity and specificity on the tuning set was minimized.

The tuning set was independently graded by three glaucoma specialists. Since this set was not adjudicated, a majority vote was used.

| | | Round 1 | | | | Round 2 (Reference Standard) | | | |
|---|---|---|---|---|---|---|---|---|---|
| | | Non-glaucoma | Low-risk | High-risk | Likely | Non-glaucoma | Low-risk | High-risk | Likely |
| **Fully adjudicated values** | Non-glaucoma | 37 | 0 | 0 | 0 | 36 | 0 | 0 | 0 |
| | Low-risk | **6** | 26 | 0 | 0 | **2** | 29 | 0 | 0 |
| | High-risk | **1** | **1** | 14 | 0 | 0 | **1** | 13 | 0 |
| | Likely glaucoma | 0 | **1** | 0 | 4 | 0 | 0 | 0 | 5 |

Table S2. Comparison of Round 1, Round 2 (Reference Standard), and Fully Adjudicated Grades in a Set of 100 Images

A set of 100 images was fully adjudicated. Compared to the round 1 "median" (see Methods), 9 images had altered grades (in bold), changing referral decisions for 3 (bolded and underlined). Compared to the round 2 median, 3 images had altered grades (in bold), changing referral decisions for only 1 of the 100 images (bolded and underlined). Thus the round 2 median was used as the reference standard.

| Table S3. Intergrader agreement on round 1 and round 2 as measured against the Reference Standard in a set of 100 images | | |
|---|---|---|
| | Krippendorff's alpha values* | |
| Question | Round 1 | Round 2 |
| Referable GON risk | 0.89 | 0.98 |
| Glaucoma gradability | 0.87 | 0.98 |
| Rim width comparison - inferior vs. superior | 0.84 | 0.88 |
| Rim width comparison - superior vs. temporal | 0.76 | 0.94 |
| Rim notching | 0.66 | 0.76 |
| RNFL defect | 0.94 | 0.94 |
| Disc hemorrhage | 0.77 | 0.90 |
| Laminar dot sign | 0.94 | 1.0 |
| Nasalization of central vascular trunk | 0.85 | 0.95 |
| Baring of circumlinear vessels | 0.92 | 0.93 |

Abbreviations: GON, Glaucomatous Optic Neuropathy; RNFL, Retinal Nerve Fiber Layer;

* good agreement for Krippendorff's alpha (Landis and Koch)

0 to .2 "slight"; 0.21 to 0.40 "fair"; 0.41 to 0.60 "moderate"; 0.61 - 0.80 "substantial"; 0.81 to 1 "near perfect"

| Table S4. Sensitivity and Specificity of Graders Versus Algorithm Performance on Validation Dataset "A" Subset (n=411) | | | | |
|---|---|---|---|---|
| Grader/Algorithm | Sensitivity [95% CI] | p-value | Specificity [95% CI] | p-value |
| Algorithm (Balanced Operating Point) | 0.761 [0.645 - 0.853] | Reference point | 0.923 [0.889 - 0.949] | Reference point |
| Algorithm (High Sensitivity Operating Point) | 0.845 [0.741 - 0.920] | N/A | 0.848 [0.806 - 0.885] | N/A |
| Algorithm (High Specificity Operating Point) | 0.662 [0.541 - 0.769] | N/A | 0.966 [0.941 - 0.983] | N/A |
| Glaucoma Specialist 1 | 0.639 [0.517 - 0.749] | 0.108 | 0.855 [0.813 - 0.891] | 0.024 |
| Glaucoma Specialist 2 | 0.528 [0.407 - 0.647] | 0.002 | 0.882 [0.843 - 0.914] | 0.302 |
| Glaucoma Specialist 3 | 0.375 [0.264 - 0.497] | < 0.001 | 0.906 [0.869 - 0.935] | 1 |
| Ophthalmologist 1 | 0.625 [0.503 - 0.736] | 0.041 | 0.855 [0.813 - 0.891] | 0.008 |
| Ophthalmologist 2 | 0.292 [0.190 - 0.411] | < 0.001 | 0.906 [0.869 - 0.935] | 1 |
| Ophthalmologist 3 | 0.583 [0.461 - 0.698] | 0.004 | 0.885 [0.846 - 0.917] | 0.349 |
| Ophthalmologist 4 | 0.736 [0.619 - 0.833] | 0.804 | 0.758 [0.709 - 0.803] | <0.001 |
| Optometrist 1 | 0.514 [0.393 - 0.633] | < 0.001 | 0.882 [0.843 - 0.914] | 0.291 |
| Optometrist 2 | 0.708 [0.589 - 0.810] | 0.455 | 0.876 [0.836 - 0.909] | 0.154 |
| Optometrist 3 | 0.292 [0.190 - 0.411] | < 0.001 | 0.926 [0.893 - 0.952] | 0.324 |

Abbreviations: CI, Confidence interval;

| Table S5. Sensitivity And Specificity of Graders Versus Algorithm Performance on validation dataset "A" Subsets Deemed Gradable (out of n=411)* ||||||||
|---|---|---|---|---|---|---|---|
| Grader | Number deemed gradable | Sensitivity [95% CI] | Sensitivity of Algorithm** [95% CI] | p-value* | Specificity [95% CI] | Specificity of Algorithm** [95% CI] | p-value** |
| Glaucoma Specialist 1 | 391 | 0.639 [0.517 - 0.749] | 0.845 [0.740 - 0.920] | 0.006 | 0.912 [0.875 - 0.941] | 0.855 [0.812 - 0.892] | 0.022 |
| Glaucoma Specialist 2 | 386 | 0.528 [0.407 - 0.647] | 0.845 [0.740 - 0.920] | < 0.001 | 0.955 [0.926 - 0.975] | 0.856 [0.812 - 0.893] | < 0.001 |
| Glaucoma Specialist 3 | 397 | 0.375 [0.264 - 0.497] | 0.845 [0.740 - 0.920] | < 0.001 | 0.948 [0.917 - 0.969] | 0.852 [0.808 - 0.889] | < 0.001 |
| Ophthalmologist 1 | 398 | 0.625 [0.503 - 0.736] | 0.845 [0.740 - 0.920] | < 0.001 | 0.892 [0.853 - 0.924] | 0.846 [0.802 - 0.884] | 0.036 |
| Ophthalmologist 2 | 392 | 0.292 [0.190 - 0.411] | 0.845 [0.740 - 0.920] | < 0.001 | 0.962 [0.935 - 0.980] | 0.846 [0.802 - 0.884] | < 0.001 |
| Ophthalmologist 3 | 382 | 0.571 [0.447 - 0.689] | 0.855 [0.750 - 0.928] | < 0.001 | 0.965 [0.938 - 0.982] | 0.855 [0.811 - 0.892] | < 0.001 |
| Ophthalmologist 4 | 389 | 0.729 [0.609 - 0.828] | 0.870 [0.767 - 0.939] | 0.013 | 0.808 [0.761 - 0.850] | 0.852 [0.808 - 0.889] | 0.070 |
| Optometrist 1 | 395 | 0.514 [0.393 - 0.633] | 0.845 [0.740 - 0.920] | < 0.001 | 0.929 [0.895 - 0.954] | 0.851 [0.807 - 0.888] | 0.0006 |
| Optometrist 2 | 390 | 0.704 [0.584 - 0.807] | 0.857 [0.753 - 0.929] | 0.003 | 0.934 [0.901 - 0.959] | 0.846 [0.801 - 0.884] | < 0.001 |
| Optometrist 3 | 386 | 0.250 [0.153 - 0.370] | 0.896 [0.797 - 0.957] | < 0.001 | 0.991 [0.973 - 0.998] | 0.858 [0.815 - 0.895] | < 0.001 |

Abbreviations: CI, Confidence interval;

* Sensitivity and specificity of each grader were calculated based on the subset of images that grader deemed gradable

**Algorithm was evaluated at the balanced operating point on the subset graders deemed gradable, and p-value is a comparison to this point

| | Train Set (N=44,244) | | | Tuning Set (N=1,203) | | |
|---|---|---|---|---|---|---|
| | Odds Ratio | p-value | Rank | Odds Ratio | p-value | Rank |
| Vertical CD Ratio ≥ 0.7 | 43.980 | < 0.001 | 1 | 183.732 | < 0.001 | 1 |
| Rim Comparison: S < T | 2.937 | < 0.001 | 2 | 3.177 | 0.153 | 7 |
| Circumlinear Vessels: Present + Bared | 2.639 | < 0.001 | 3 | 8.087 | < 0.001 | 3 |
| RNFL Defect: Possible or Yes | 2.258 | < 0.001 | 4 | 8.908 | < 0.001 | 2 |
| Laminar Dot: Possible or Yes | 2.064 | < 0.001 | 5 | 3.386 | < 0.001 | 6 |
| Rim Comparison: I < S | 2.029 | < 0.001 | 6 | 1.601 | 0.153 | 11 |
| Notch: Possible or Yes | 1.573 | < 0.001 | 7 | 3.791 | 0.180 | 4 |
| Nasalization Directed: Yes | 1.398 | < 0.001 | 8 | 2.171 | < 0.001 | 9 |
| Nasalization Emerging: Yes | 0.935 | 0.044 | 9 | 3.667 | < 0.001 | 5 |
| Beta PPA: Possible or Yes | 0.876 | < 0.001 | 10 | 2.227 | < 0.001 | 8 |
| Disc Hemorrhage: Possible or Yes | 0.269 | < 0.001 | 11 | 1.796 | 0.456 | 10 |

Table S6. Logistic Regression Models to Understand the Relative Importance of Individual Optic Nerve Head Features for Glaucoma Referral Decisions in the development sets

Abbreviations: Sig., Significance; CD, Cup-to-disc; RNFL, Retinal Nerve Fiber Layer; S, Superior; T, Temporal; I, Inferior; PPA, ParaPapillary Atrophy;

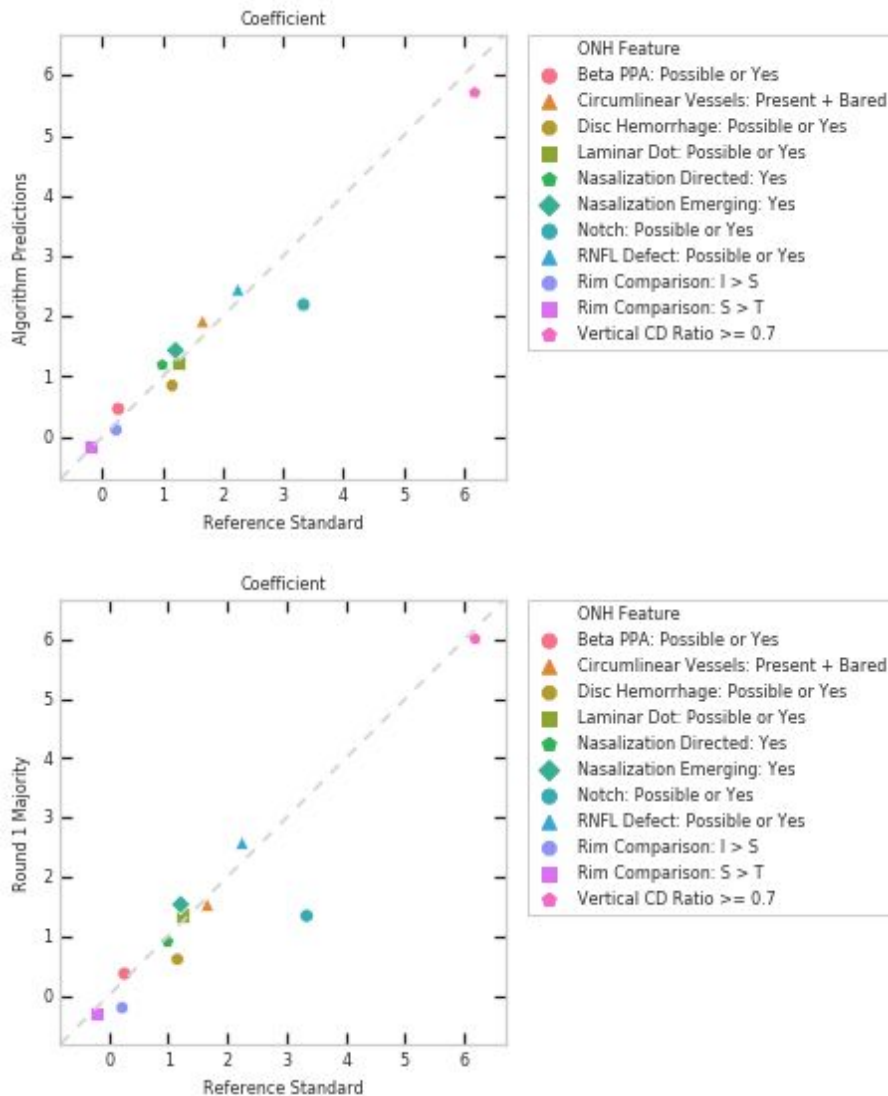

**Figure S1A-B**: Comparison of the relative importance of ONH features for refer/no-refer decisions in validation dataset A for **(A)** Algorithm Predictions vs. Reference Standard and **(B)** Round 1 Majority vs. Reference Standard. Feature importance is quantified as the beta coefficient, the natural log of the odd's ratio from the corresponding logistic regression (**Table 2**).

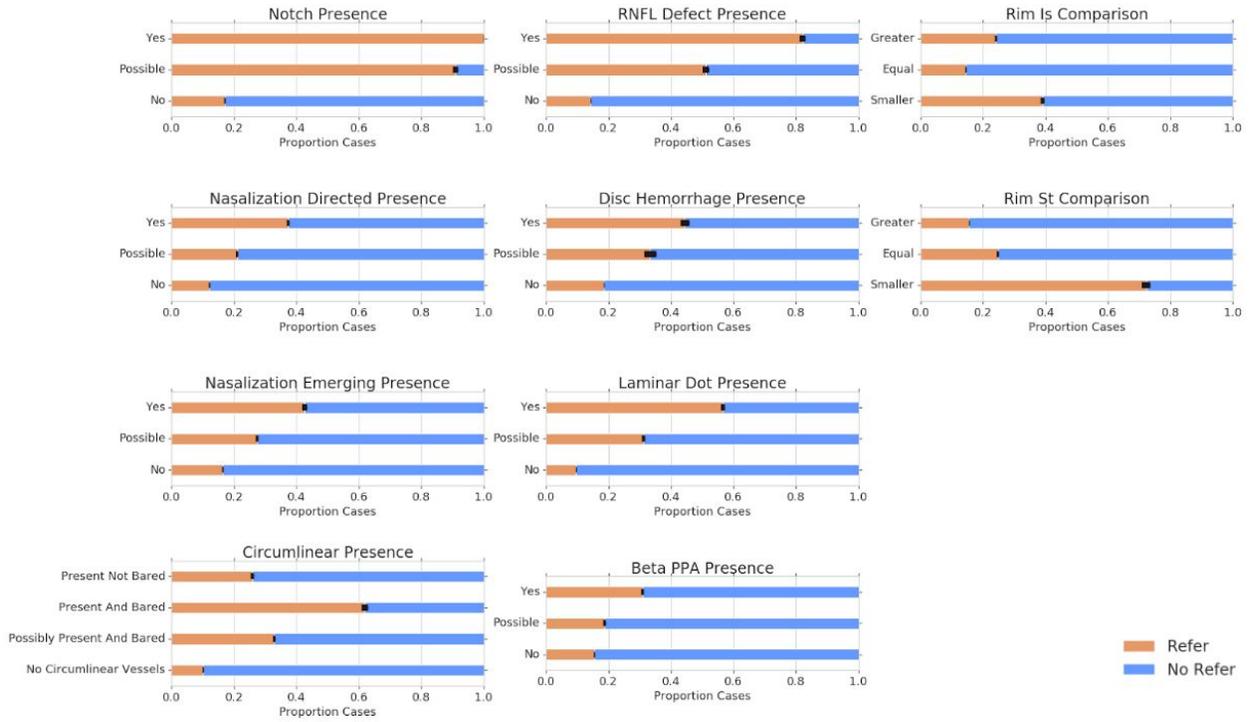

**Figure S2**: Proportions of all ONH feature grades amongst refer/no-refer from validation dataset "A". Error bars represent 95% confidence intervals.
Abbreviations: ONH, Optic Nerve Head.

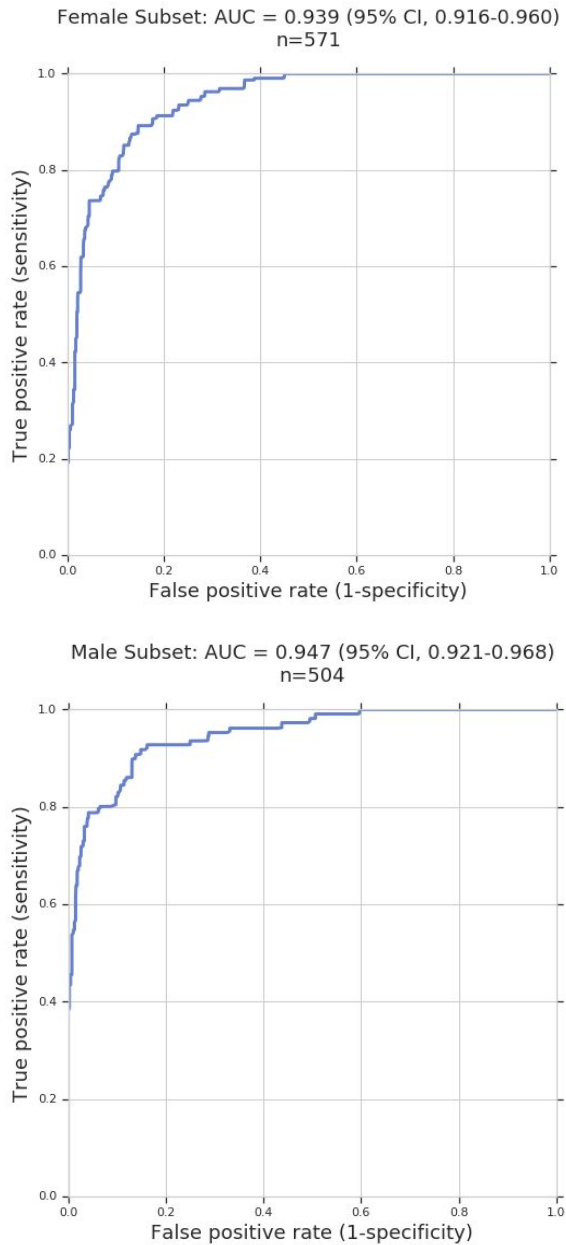

**Figure S3A-B**: Receiver operating characteristic (ROC) curve analyses for referable Glaucoma in subsets of validation dataset "A" with only females (n=571) **(A)** and only males (n=504) **(B)**. Self-reported sex was only available for 1,075 images.

## References

1. Quigley HA, Broman AT. The number of people with glaucoma worldwide in 2010 and 2020. *Br J Ophthalmol*. 2006;90(3):262-267.

2. Tham Y-C, Li X, Wong TY, Quigley HA, Aung T, Cheng C-Y. Global prevalence of glaucoma and projections of glaucoma burden through 2040: a systematic review and meta-analysis. *Ophthalmology*. 2014;121(11):2081-2090.

3. Leite MT, Sakata LM, Medeiros FA. Managing glaucoma in developing countries. *Arq Bras Oftalmol*. 2011;74(2):83-84.

4. Rotchford AP, Kirwan JF, Muller MA, Johnson GJ, Roux P. Temba glaucoma study: a population-based cross-sectional survey in urban South Africa. *Ophthalmology*. 2003;110(2):376-382.

5. Hennis A, Wu S-Y, Nemesure B, Honkanen R, Cristina Leske M. Awareness of Incident Open-angle Glaucoma in a Population Study. *Ophthalmology*. 2007;114(10):1816-1821.

6. Prum BE Jr, Lim MC, Mansberger SL, et al. Primary Open-Angle Glaucoma Suspect Preferred Practice Pattern(®) Guidelines. *Ophthalmology*. 2016;123(1):P112-P151.

7. Weinreb RN. *Glaucoma Screening*. Kugler Publications; 2008.

8. Newman-Casey PA, Verkade AJ, Oren G, Robin AL. Gaps in Glaucoma care: A systematic review of monoscopic disc photos to screen for glaucoma. *Expert Rev Ophthalmol*. 2014;9(6):467-474.

9. Bernardes R, Serranho P, Lobo C. Digital ocular fundus imaging: a review. *Ophthalmologica*. 2011;226(4):161-181.

10. Shi L, Wu H, Dong J, Jiang K, Lu X, Shi J. Telemedicine for detecting diabetic retinopathy: a systematic review and meta-analysis. *Br J Ophthalmol*. 2015;99(6):823-831.

11. Weinreb RN, Khaw PT. Primary open-angle glaucoma. *Lancet*. 2004;363(9422):1711-1720.

12. Bowd C, Weinreb RN, Zangwill LM. Evaluating the optic disc and retinal nerve fiber layer in glaucoma. I: Clinical examination and photographic methods. *Semin Ophthalmol*. 2000;15(4):194-205.

13. Weinreb RN, Aung T, Medeiros FA. The Pathophysiology and Treatment of Glaucoma. *JAMA*. 2014;311(18):1901.

14. Prum BE Jr, Rosenberg LF, Gedde SJ, et al. Primary Open-Angle Glaucoma Preferred Practice Pattern(®) Guidelines. *Ophthalmology*. 2016;123(1):P41-P111.

15. Hollands H, Johnson D, Hollands S, Simel DL, Jinapriya D, Sharma S. Do Findings on Routine Examination Identify Patients at Risk for Primary Open-Angle Glaucoma? *JAMA*. 2013;309(19):2035.

16. Mardin CY, Horn F, Viestenz A, Lämmer R, Jünemann A. [Healthy optic discs with large cups--a diagnostic challenge in glaucoma]. *Klin Monbl Augenheilkd*. 2006;223(4):308-314.

17. Jonas JB, Fernández MC. Shape of the neuroretinal rim and position of the central retinal vessels in glaucoma. *Br J Ophthalmol*. 1994;78(2):99-102.

18. Jonas JB, Schiro D. Localized retinal nerve fiber layer defects in nonglaucomatous optic nerve atrophy. *Graefes Arch Clin Exp Ophthalmol*. 1994;232(12):759-760.

19. Chihara E, Matsuoka T, Ogura Y, Matsumura M. Retinal nerve fiber layer defect as an early manifestation of diabetic retinopathy. *Ophthalmology*. 1993;100(8):1147-1151.

20. Chaum E, Drewry RD, Ware GT, Charles S. Nerve fiber bundle visual field defect resulting from a giant peripapillary cotton-wool spot. *J Neuroophthalmol*. 2001;21(4):276-277.

21. Sutton GE, Motolko MA, Phelps CD. Baring of a circumlinear vessel in glaucoma. *Arch Ophthalmol*. 1983;101(5):739-744.

22. Fingeret M, Medeiros FA, Susanna R Jr, Weinreb RN. Five rules to evaluate the optic disc and retinal nerve fiber layer for glaucoma. *Optometry*. 2005;76(11):661-668.

23. LeCun Y, Bengio Y, Hinton G. Deep learning. *Nature*. 2015;521(7553):436-444.

24. Gulshan V, Peng L, Coram M, et al. Development and Validation of a Deep Learning Algorithm for Detection of Diabetic Retinopathy in Retinal Fundus Photographs. *JAMA*. 2016;316(22):2402-2410.

25. Krause J, Gulshan V, Rahimy E, et al. Grader Variability and the Importance of Reference Standards for Evaluating Machine Learning Models for Diabetic Retinopathy. *Ophthalmology*. 2018;125(8):1264-1272.

26. Ting DSW, Cheung CY-L, Lim G, et al. Development and Validation of a Deep Learning System for Diabetic Retinopathy and Related Eye Diseases Using Retinal Images From Multiethnic Populations With Diabetes. *JAMA*. 2017;318(22):2211-2223.

27. Sayres R, Taly A, Rahimy E, et al. Using a deep learning algorithm and integrated gradients explanation to assist grading for diabetic retinopathy. *Ophthalmology*. November 2018. doi:10.1016/j.ophtha.2018.11.016

28. Liu S, Graham SL, Schulz A, et al. A Deep Learning-Based Algorithm Identifies Glaucomatous Discs Using Monoscopic Fundus Photographs. *Ophthalmology Glaucoma*. 2018;1(1):15-22.

29. Li Z, He Y, Keel S, Meng W, Chang RT, He M. Efficacy of a Deep Learning System for Detecting Glaucomatous Optic Neuropathy Based on Color Fundus Photographs. *Ophthalmology*. 2018;125(8):1199-1206.

30. Shibata N, Tanito M, Mitsuhashi K, et al. Development of a deep residual learning algorithm to screen for glaucoma from fundus photography. *Sci Rep*. 2018;8(1):14665.

31. Christopher M, Belghith A, Bowd C, et al. Performance of Deep Learning Architectures and Transfer Learning for Detecting Glaucomatous Optic Neuropathy in Fundus Photographs. *Sci Rep*. 2018;8(1):16685.

32. Thomas S-M, Jeyaraman MM, Hodge WG, Hutnik C, Costella J, Malvankar-Mehta MS. The effectiveness of teleglaucoma versus in-patient examination for glaucoma screening: a systematic review and meta-analysis. *PLoS One*. 2014;9(12):e113779.

33. Welcome to EyePACS. http://www.eyepacs.org. Accessed December 5, 2018.

34. [No title]. http://www.inoveon.com/. Accessed December 5, 2018.

35. Age-Related Eye Disease Study Research Group. The Age-Related Eye Disease Study (AREDS): design implications. AREDS report no. 1. *Control Clin Trials*. 1999;20(6):573-600.

36. Website. http://www.ukbio- bank.ac.uk/about-biobank-uk. Accessed December 5, 2018.

37. Lopes FSS, Dorairaj S, Junqueira DLM, Furlanetto RL, Biteli LG, Prata TS. Analysis of neuroretinal rim distribution and vascular pattern in eyes with presumed large physiological cupping: a comparative study. *BMC Ophthalmol*. 2014;14:72.

38. Susanna R Jr. The lamina cribrosa and visual field defects in open-angle glaucoma. *Can J Ophthalmol*. 1983;18(3):124-126.

39. Poon LY-C, Valle DS-D, Turalba AV, et al. The ISNT Rule: How Often Does It Apply to Disc Photographs and Retinal Nerve Fiber Layer Measurements in the Normal Population? *Am J Ophthalmol*. 2017;184:19-27.

40. Szegedy C, Vanhouke V, Ioffe S, Shlens J, Wojna Z. Rethinking the Inception Architecture for Computer Vision. *arXiv preprint arXiv:150203167*. December 2015. http://arxiv.org/pdf/1512.00567v3.pdf.

41. TensorFlow - BibTex Citation. https://chromium.googlesource.com/external/github.com/tensorflow/tensorflow/+/0.6.0/tensorflow/g3doc/resources/bib.md. Accessed December 3, 2018.

42. Krizhevsky A, Sutskever I, Hinton GE. Imagenet classification with deep convolutional neural networks. In: *Advances in Neural Information Processing Systems.* ; 2012:1097-1105.

43. Settles B. Active Learning. *Synthesis Lectures on Artificial Intelligence and Machine Learning*. 2012;6(1):1-114.

44. Opitz D, Maclin R. Popular Ensemble Methods: An Empirical Study. *J Artif Intell Res*. 1999;11:169-198.

45. Chihara LM, Hesterberg TC. *Mathematical Statistics with Resampling and R*.; 2018.

46. Clopper CJ, Pearson ES. The Use of Confidence or Fiducial Limits Illustrated in the Case of the Binomial. *Biometrika*. 1934;26(4):404.

47. Massey FJ. The Kolmogorov-Smirnov Test for Goodness of Fit. *J Am Stat Assoc*. 1951;46(253):68.

48. Krippendorff K. *Content Analysis: An Introduction to Its Methodology*. SAGE Publications; 2018.

49. Herschler J, Osher RH. Baring of the circumlinear vessel. An early sign of optic nerve damage. *Arch Ophthalmol*. 1980;98(5):865-869.

50. Susanna R Jr, Medeiros FA. *The Optic Nerve in Glaucoma*.; 2006.

51. Tielsch JM, Katz J, Quigley HA, Miller NR, Sommer A. Intraobserver and interobserver agreement in measurement of optic disc characteristics. *Ophthalmology*. 1988;95(3):350-356.

52. Varma R, Steinmann WC, Scott IU. Expert agreement in evaluating the optic disc for glaucoma. *Ophthalmology*. 1992;99(2):215-221.

53. Zangwill LM, Weinreb RN, Berry CC, et al. Racial differences in optic disc topography: baseline results from the confocal scanning laser ophthalmoscopy ancillary study to the ocular hypertension treatment study. *Arch Ophthalmol*. 2004;122(1):22-28.

54. Lee RY, Kao AA, Kasuga T, et al. Ethnic variation in optic disc size by fundus photography. *Curr Eye Res*. 2013;38(11):1142-1147.

55. Tatham AJ, Medeiros FA. Detecting Structural Progression in Glaucoma with Optical Coherence Tomography. *Ophthalmology*. 2017;124(12S):S57-S65.

56. Muhammad H, Fuchs TJ, De Cuir N, et al. Hybrid Deep Learning on Single Wide-field Optical Coherence tomography Scans Accurately Classifies Glaucoma Suspects. *J Glaucoma*. 2017;26(12):1086-1094.

57. Asaoka R, Murata H, Hirasawa K, et al. Using Deep Learning and Transfer Learning to Accurately Diagnose Early-Onset Glaucoma From Macular Optical Coherence Tomography Images. *American Journal of Ophthalmology*. 2019;198:136-145. doi:10.1016/j.ajo.2018.10.007